\pdfoutput=1
\documentclass[11pt]{article}

\usepackage[preprint]{acl}

\usepackage{times}
\usepackage{latexsym}

\usepackage[T1]{fontenc}
\usepackage[utf8]{inputenc}

\usepackage{microtype}

\usepackage{inconsolata}

\usepackage{graphicx}

\usepackage{tabularx}
\usepackage{booktabs}

\usepackage{algorithm}
\usepackage[noend]{algpseudocode}
\usepackage[normalem]{ulem}

\usepackage{amsfonts}

\usepackage{todonotes}
\usepackage{amsmath}
\usepackage{xspace}
\usepackage{listings} % for lstlisting
\usepackage{amsthm}
\usepackage{centernot}
\usepackage{hyperref}
\usepackage{mathtools}
\usepackage[most]{tcolorbox}

\theoremstyle{definition}

\newcommand{\NAME}{HyGenar\xspace} 

\definecolor{red}{RGB}{220, 50, 47}
\definecolor{promptbg}{RGB}{240, 248, 255} 
\definecolor{feedbackbg}{RGB}{255, 204, 204}

\newcounter{promptblockcounter}
\renewcommand{\thepromptblockcounter}{\arabic{promptblockcounter}}
\newenvironment{promptblock}[2][htbp]
{
    \refstepcounter{promptblockcounter}
    \begin{tcolorbox}[
        float, 
        floatplacement=#1,
        colback=promptbg,
        colframe=blue!50!black,
        boxrule=0.75pt, 
        arc=3pt, 
        title={Prompt Template~\thepromptblockcounter: #2}, 
        before upper=\raggedright,
    ]
}
{
    \end{tcolorbox}
}

\newcounter{feedbackblockcounter}
\renewcommand{\thefeedbackblockcounter}{\arabic{feedbackblockcounter}}
\newenvironment{feedbackblock}[2][htbp]
{
    \refstepcounter{feedbackblockcounter}
    \begin{tcolorbox}[
        float, 
        floatplacement=#1,
        colback=feedbackbg,
        colframe=red!70!black,
        boxrule=0.75pt, 
        arc=3pt, 
        title={Parser Feedback ~\thefeedbackblockcounter: #2}, 
        before upper=\raggedright,
    ]
}
{
    \end{tcolorbox}
}

\newcommand{\bnf}{G}
\newcommand{\guess}{\bnf^*}
\newcommand{\bnfref}{\bnf^{\mathit{ref}}}
\newcommand{\cfg}{\mathcal{G}}
\newcommand{\languageOf}[1]{\ensuremath{\mathcal{L}(#1)}}
\newcommand{\diff}{{\mathit{Diff}}}
\newcommand{\of}{\mathit{OF}}
\newcommand{\og}{\mathit{OG}}
\newcommand{\sx}{\mathit{SX}}
\newcommand{\se}{\mathit{SE}}
\newcommand{\tu}{\mathit{TU}}
\newcommand{\grammarset}{\mathbb{G}}

\newenvironment{squishitem}{
\begin{itemize}
   \setlength{\itemsep}{1pt}
   \setlength{\parskip}{0pt}
   \setlength{\parsep}{0pt}
}
{\end{itemize}}

\title{\NAME: An LLM-Driven Hybrid Genetic Algorithm for Few-Shot Grammar Generation}
\author{
 \textbf{Weizhi Tang},
 \textbf{Yixuan Li},
 \textbf{Chris Sypherd},
 \textbf{Elizabeth Polgreen},
 \textbf{Vaishak Belle}
\\
 University of Edinburgh
\\
 \small{
   \href{mailto:email@domain}{Weizhi.Tang@ed.ac.uk}
 },
 \small{
   \href{mailto:email@domain}{yixuan.li.cs@ed.ac.uk}
 },
 \small{
   \href{mailto:email@domain}{c.n.sypherd@sms.ed.ac.uk}
 },\\
 \small{
   \href{mailto:email@domain}{elizabeth.polgreen@ed.ac.uk}
 },
 \small{
   \href{mailto:email@domain}{vbelle@ed.ac.uk}
 }
}

\begin{document}
\maketitle
\begin{abstract}
Grammar plays a critical role in natural language processing and text/code generation by enabling the definition of syntax, the creation of parsers, and guiding structured outputs. Although large language models (LLMs) demonstrate impressive capabilities across domains, their ability to infer and generate grammars has not yet been thoroughly explored. In this paper, we aim to study and improve the ability of LLMs for few-shot grammar generation, where grammars are inferred from sets of a small number of positive and negative examples and generated in Backus-Naur Form. To explore this, we introduced a novel dataset comprising 540 structured grammar generation challenges, devised 6 metrics, and evaluated 8 various LLMs against it. Our findings reveal that existing LLMs perform sub-optimally in grammar generation. To address this, we propose an LLM-driven hybrid genetic algorithm, namely \NAME, to optimize grammar generation. \NAME achieves substantial improvements in both the syntactic and semantic correctness of generated grammars across LLMs~\footnote{The code is open-source and available at \url{https://github.com/RutaTang/HyGenar}.}.
\end{abstract}

% Introduction
\section{Introduction}

Grammar inference, also known as grammar induction, consists of inferring a grammar from a set of examples~\cite{horning1969study,de2010grammatical,Stevenson_Cordy_2014,D_Ulizia_Ferri_Grifoni_2011}. It has been studied and used in various fields, such as natural language processing, where it can reduce the effort required to generate syntactic or semantic models automatically~\cite{kai2024leveraginggrammarinductionlanguage, Dulizia2011A}, and software engineering, where inferred grammars can guide reverse engineering and automated parser generation~\cite{Stevenson2014A}. By relying on characteristic examples~\cite{de2010grammatical}, grammar inference enables automated discovery of underlying syntactic or structural patterns. 

\begin{figure}[t]
  \includegraphics[width=\columnwidth]{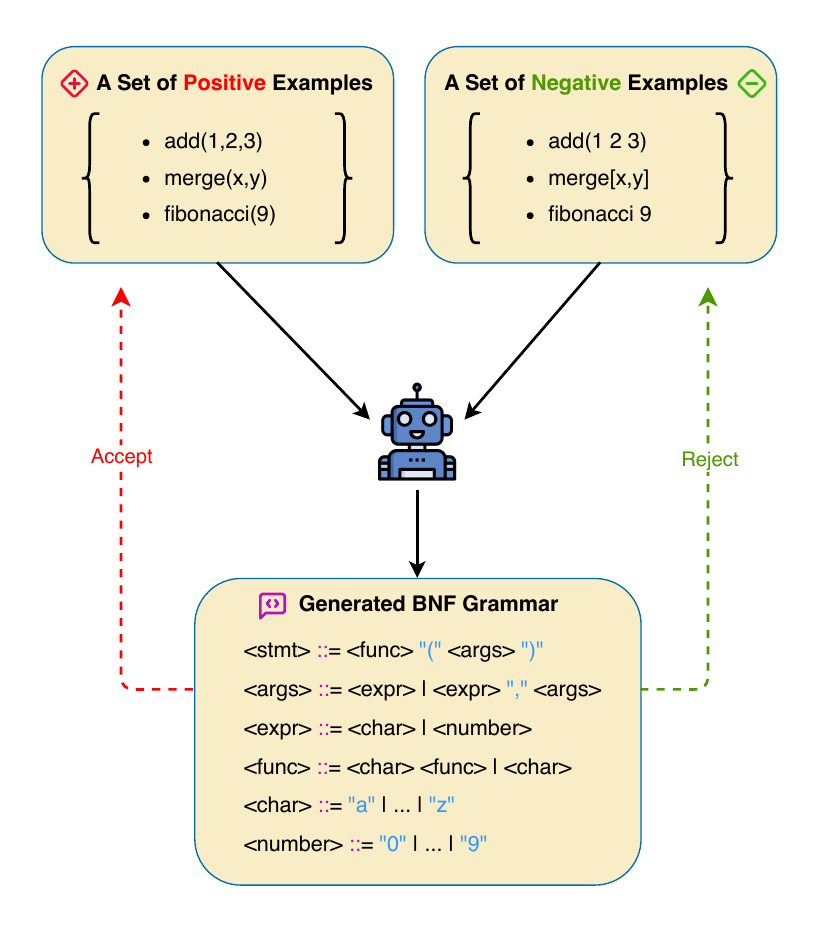}
  \caption{Given a small set of positive and negative examples, LLMs should infer and generate a grammar that accepts all positives and rejects all negatives.}
  \label{fig:bnf_generation}
\end{figure}

Backus-Naur Form (BNF) is used to define the grammar of formal languages~\cite{Chomsky_1956,backus1959syntax,backus1960report}, typically Context-Free Grammars (CGFs)~\cite{Chomsky_1956, Aho_Aho_2007}. It has also been used in various ways such as in parser generators like ANTLR4~\cite{ANTLR4} and Yacc~\cite{Johnson1978YaccYA} or in constraining output structure of Large Language Models (LLMs)~\cite{willard2023efficientguidedgenerationlarge,BeurerKellner2024GuidingLT}.

Although LLMs have exhibited remarkable capabilities across diverse domains~\cite{gaur-saunshi-2023-reasoning,imani-etal-2023-mathprompter,Pan_Albalak_Wang_Wang_2023,tang2024tom,LLM-SYGUS,jiang2024surveylargelanguagemodels}, their capacity for grammar inference, and particularly for generating grammars in BNF, has not yet been well explored. This paper focuses on investigating and improving the ability of LLMs to infer a CFG and generate it in BNF from a given set of positive and negative examples. A correctly generated grammar should accept all positives and reject all negatives. Typically, grammar inference requires a large set of characteristic examples that can uniquely identify a CFG~\cite{de2010grammatical}. However, we instead explore whether LLMs can infer a CFG from fewer examples without imposing any constraints on the examples, based on their experience and knowledge acquired through training on large-scale corpora. We refer to this process as ``few-shot grammar generation'', emphasizing both the grammar inference of CFGs with fewer examples and their generation in BNF, and we use ``grammar'' to denote a CFG represented in BNF. An example is shown in Figure~\ref{fig:bnf_generation}.

To explore the capacity of LLMs for few-shot grammar generation, we first construct a dedicated dataset of 540 challenges, each with only 3 positive and 3 negative examples, and use it to evaluate the performance of 8 LLMs, encompassing both open- and closed-source models of varying parameter sizes, including those specialized for code generation. We then propose an LLM-driven hybrid genetic algorithm, namely \NAME, which adapts the principles of genetic algorithms with the integration of LLM-driven population initialization and mutation. We devised and adopted 6 metrics to comprehensively evaluate and analyze their performance from the perspectives of syntax and semantic correctness, over-fitting, over-generalization, and utility. The results show that, while most LLMs demonstrate unsatisfactory performance, our proposed algorithm significantly enhances the grammar generation ability across most of the evaluated LLMs.

We summarize the main contributions of this paper as follows:
\begin{enumerate}
    \item We constructed a dedicated dataset of 540 challenges for few-shot grammar generation and comprehensively evaluated 8 LLMs;
    \item We designed 6 metrics for measuring the ability of LLMs in this task and performed an extensive analysis based on them; 
    \item We proposed a novel method, \NAME, to improve the grammar generation performance of LLMs and showed that it achieves significant improvements across LLMs.
\end{enumerate}

%Motivation
% \input{sections/motivation}

% Background
\section{Background}
\subsection{Context-Free Grammar}

A \emph{context-free grammar} (CFG) consists of terminals, non-terminals, a start symbol, and production rules~\cite{hopcroft2001introduction, Aho_Aho_2007}. It can be formally defined as a quadruple $\cfg = (V, \Sigma, \Pi, S)$, where $V$ is a finite set of non-terminal symbols, $\Sigma$ is the set of terminals, $\Sigma \cap V = \emptyset$, $\Pi$ is a finite set of production rules, $\Pi \subseteq V \times (V \cup \Sigma)^\ast$, and $S \in V$ is the start symbol of the grammar.
Elements of $(V \cup \Sigma)^*$ are known as sentential forms.

The language generated by $\cfg$, denoted $\languageOf{\cfg}$, comprises all strings derivable from $S$ using the rules in $\Pi$: 
$
\languageOf{\cfg} = \{ \sigma \in \Sigma^\ast \mid S \stackrel{*}{\rightarrow} \sigma \}
$. For $\alpha,\beta \in (V\cup \Sigma)^*$, we say $\alpha$ directly derives $\beta$ in one step as $\alpha {\rightarrow} \beta$, and define $\alpha \stackrel{*}{\rightarrow} \beta$ as $\alpha$ deriving $\beta$ in zero or more steps if there exists a finite sequence of $\gamma_0,\dots,\gamma_n \in (V \cup \Sigma)^*$ where $n\geq 0$, such that $\alpha = \gamma_0 \rightarrow \gamma_1 \rightarrow \dots \rightarrow \gamma_n = \beta$.

\subsection{Backus-Naur Form}
\label{sec:bnf_background}
Backus–Naur Form (BNF) is a notation used to define the grammar of formal languages, typically CFG~\cite{Chomsky_1956,backus1959syntax,backus1960report}.  
In this paper, LLM generates grammars in BNF. 
% In BNF, production rules are defined  

A context-free grammar $\cfg = (V, \Sigma, \Pi, S)$ is given in BNF notation as a list of grouped production rules, where each rule group is written:

$$\texttt{<$v_i$>}\vcentcolon\vcentcolon=\alpha_1 | \alpha_2 | \ldots$$

where $v_i$ enclosed between the pair ``\texttt{< >}" is a nonterminal symbol in $V$, and $\alpha_1, \alpha_2 \ldots \in (\Sigma \cup V)^*$ is the list of sentential forms that can be derived in one step from $v_i$. This represents the set of all production rules with $v_i$ on the left-hand side, i.e., $v_i \rightarrow \alpha_1, v_i \rightarrow \alpha_2, \ldots$. 

We extend the definition of a CFG to define a grammar in BNF to be $\bnf = (V, \Sigma, \Pi, S, R)$, where $V, \Sigma, \Pi$, and $S$ are defined as before. $R$ is a set of sets of production rules, denoted $\{r_1, \ldots r_n\}$. Each $r_i \in R$ is a set of production rules where the left-hand side of all rules is the $i^{th}$ non-terminal symbol $v_i$, i.e., $r_i$ is $(v_i \times (\Sigma \cup V)^*) \cap \Pi$. 
Since each non-terminal symbol has a corresponding rule set, $n=|V|=|R|$. The language of the BNF grammar $\bnf$ is denoted by $\languageOf{\bnf}$ and is equal to the language of the original CFG, $\languageOf{\cfg}$.

We say that a grammar $\bnf$ is a \emph{valid} grammar, and that $valid(\bnf)$ evaluates to true, if it has a correct BNF syntax, and if $R$ is as defined above, and if all nonterminal symbols have at least one rule in their corresponding rule set.

\subsection{Grammar Inference}
Grammar inference aims to learn grammar automatically from a set of examples~\cite{horning1969study,de2010grammatical,Stevenson_Cordy_2014,D_Ulizia_Ferri_Grifoni_2011}. In this paper, we focus on inferring a CFG, in BNF notation, from a small set of positive and negative examples. 

Given a set of positive examples $\mathcal{P}$ and negative examples $\mathcal{N}$, consisting of strings that must be, respectively, accepted and rejected, the objective is to infer a BNF grammar $\bnf$. The generated $\bnf$ should satisfy $\mathcal{P} \subseteq \languageOf{\bnf}$ and $\mathcal{N} \cap \languageOf{\bnf} = \emptyset$, which ensures $\bnf$ accepts all positive examples and rejects all negative examples.

% \subsection{Genetic Algorithm}
% Genetic Algorithm (GA) is a population-based metaheuristic method inspired by the principles of natural selection and genetics in biological evolution~\cite{Michalewicz_Schoenauer_1996,Katoch_Chauhan_Kumar_2021,Gendreau_Potvin_2019}. Typically, a GA relies on a set of genetic operators consisting of encoding, fitness, selection, crossover, and mutation~\cite{Katoch_Chauhan_Kumar_2021}. Based on these operators, a GA operates on a population of candidate solutions, refining the population over multiple generations, to find an optimal or near-optimal solution for a given problem.

% Related Work
\section{Related Work}
\subsection{Grammar Generation}
Grammar inference has been widely studied and applied across various fields, such as natural language processing~\cite{kai2024leveraginggrammarinductionlanguage,D_Ulizia_Ferri_Grifoni_2011}, bio-informatics~\cite{de2010grammatical}, pattern recognition~\cite{Pedro2013Using, Richetin1984Efficient,de2010grammatical}, and software engineering~\cite{Schröder_Cito_2022, Stevenson_Cordy_2014}. Previous works have also proposed various approaches for grammar inference~\cite{Rodrigues2007GeneticPF, Cohen2017InducingRG,ga-metagrammar, Dulizia2011A, chen1995bayesian}. However, few works are directly related to exploring the ability of LLMs for few-shot grammar generation, which, to reiterate, is to infer grammars from a small set of positives and negatives while generating them in BNF. 

\subsection{Code Generation}
LLMs demonstrate the ability of code generation~\cite{Jiang2024A,Huang2023AgentCoder,dehaerne2022}, with various approaches proposed to improve it~\cite{shinn2023reflexion, madaan2023selfrefine, Huang2023AgentCoder, Jiang2023SelfPlanning, Chen2023Teaching}. We consider grammar generation to share notable similarities with code generation, since in grammar generation it is not only required to infer grammars but also to generate grammars in BNF. Thus, following a similar approach to Reflexion~\cite{shinn2023reflexion} and Self-Refine~\cite{madaan2023selfrefine} to enhance code generation, we propose a method as one of the baselines for evaluation.

% BNF Generation in LLMs
\section{Grammar Generation Ability of LLMs}
In this section, we describe the dedicated dataset constructed to evaluate the ability of LLMs in grammar generation, introduce 6 metrics we use in evaluation, explain experiments, and analyze results in detail. We detail each in the following subsections.

\subsection{Dataset}
To evaluate the capacity of LLMs for grammar generation, we present a dedicated dataset. 

During dataset construction, for each $k \in \{1,2,\dots,9\}$, we prompted GPT-4o~\cite{openai2024gpt4technicalreport} to generate 10 distinct reference grammars, where each reference grammar $\bnfref$ has $k$ nonterminal symbols and thus $|R| = k$. This gives a total of $90$ reference grammars.
% consists of exactly $k$ grammar
% edge elements. 
We used $\bnfref$ to prompt GPT-4o to produce 6 different challenges with each challenge consisting of a set of positives $\mathcal{P}$ and negatives $\mathcal{N}$ where $|\mathcal{P}|=3$ and $|\mathcal{N}|=3$, in a way that $\mathcal{P} \subseteq \languageOf{\bnfref}$ and $\mathcal{N} \cap \languageOf{\bnfref} = \emptyset$. However, GPT-4o often failed to produce challenges with valid reference grammars, positives, and negatives as $k$ increased. We manually corrected erroneous reference grammars, positives, and negatives by using a BNF parser which takes a grammar and outputs whether a grammar is in valid BNF, and whether positives are accepted, and negatives are rejected\footnote{Refer to Appendix~\ref{app:dataset-construction} for the details of dataset construction.}.

Following this process, we obtained a dataset of 540 challenges, each consisting of 3 positives and 3 negatives.
Figure~\ref{fig:bnf_generation} shows an example challenge and a corresponding solution. 

% consider a challenge with $\mathcal{P}$ as:
% \begin{center}
% \begin{minipage}{0.35\linewidth}
% \begin{verbatim}
% add(1,2,3)
% merge(x,y)
% fibonacci(9)
% \end{verbatim}
% \end{minipage}
% \end{center}
% and $\mathcal{N}$ as:
% \begin{center}
% \begin{minipage}{0.35\linewidth}
% \begin{verbatim}
% add(1 2 3)
% merge[x,y]
% fibonacci 9
% \end{verbatim}
% \end{minipage}
% \end{center}
% An example valid $\bnf$ such that $\mathcal{P} \subseteq \languageOf{\bnf}$ and $\mathcal{N} \cap \languageOf{\bnf} = \emptyset$ is as follows:
% \begin{center}
% \begin{minipage}{0.92\linewidth}
% \begin{verbatim}
% <stmt> ::= <func> "(" <args> ")"
% <args> ::= <expr> | <expr> "," <args>
% <expr> ::= <char> | <number>
% <func> ::= <char> <func> | <char>
% <char> ::= "a" | ... | "z"
% <number> ::= "0" | ... | "9"
% \end{verbatim}
% \end{minipage}
% \end{center}

\subsection{Metrics}
Let $C$ be a set of $N$ challenges where each is a tuple 
$(\bnfref,\mathcal{P},\mathcal{N}, \guess)$ consisting of a reference grammar, and a set of positive examples and negative examples, respectively, and the corresponding candidate grammar $\guess$ generated by an LLM. We evaluate the quality of generated grammars using 6 key metrics:

\paragraph{Syntax Correctness ($\sx$)}
The syntax correctness metric $\sx$ quantifies the proportion of guesses that conform to the valid BNF syntax defined in Section~\ref{sec:bnf_background}.
% A grammar generated by the LLM, $\guess$, is deemed syntactically correct if and only if it is successfully parsed by a BNF parser. 
% If the grammar is syntactically correct,we construct the corresponding grammar edge set $\mathcal{E}^\mathcal{G}$; otherwise, the grammar is rejected, and we have $\mathcal{E}^\mathcal{G}=\emptyset$
We define an indicator function as:
\begin{align*}
\mathbb{I}_{SX}(\guess) &=
\begin{cases}
1 & \text{if $valid(\guess)$,}\\
0 & \text{otherwise.}
\end{cases}
\end{align*}

\noindent $SX(C)$ is defined as:
$
% \mathrm{SX}(C) &= 
\frac{1}{N}\sum_{i=1}^N \mathbb{I}_{SX}(\guess_{i}).
$

\paragraph{Semantic Correctness ($\se$)} $\se$ captures the proportion of guesses that are semantically correct.
We define an indicator function as follows, noting that if $\guess$ is not in valid BNF, then $\languageOf{\guess} = \emptyset$:
\begin{align*}
\mathbb{I}_{SE}(\guess,\mathcal{P},\mathcal{N}) &=
\begin{cases}
1 & \text{if $\mathcal{P} \subseteq \languageOf{\guess} \,\wedge$} \\
& \mathcal{N} \cap \languageOf{\guess} = \emptyset \\
0 & \text{otherwise.}
\end{cases}
\end{align*}

% \begin{align*}
% \mathbb{I}_{SE}(\guess,\mathcal{P},\mathcal{N}) &=
% \begin{cases}
% 1 & \text{if } \big(\text{($\mathcal{E}^{\mathcal{G}}\neq \phi$)}\\
% & \quad \wedge\; (\mathcal{P}_i \subseteq \mathcal{L}^\mathcal{G}) \\
% & \quad \wedge\; (\mathcal{N}_i \centernot{\cap} \mathcal{L}^\mathcal{G})\big)\text{,} \\
% 0 & \text{otherwise.}
% \end{cases}
% \end{align*}
% meaning if only if a given grammar has correct syntax and it can accept all positive examples and reject all negative examples, it has correct semantics.

\noindent$SE(C)$ is given by:
$
\frac{1}{N}\sum_{i=1}^N \mathbb{I}_{SE}(\guess_{i},\mathcal{P}_i,\mathcal{N}_i)
$.

\paragraph{Estimating Grammar Quality}
Given a set of positive and negative examples, there are many possible valid and semantically correct solutions that are undesirable. For instance, the following grammar would be an undesirable solution for Figure~\ref{fig:bnf_generation}. Here the $\languageOf{\guess} = |P|$, and is overfitted to the examples:

\vspace{1em}
\begin{minipage}{0.92\linewidth}
\begin{verbatim}

<stmt> ::= "add(1,2,3)" | "merge(x,y)" |
           "fibonacci(9)"
\end{verbatim}
\end{minipage}
\vspace{1em}

Equally undesirable is a grammar that over-generalizes from the examples, and defines a significantly larger language than the reference grammar, because this is highly likely to contain invalid strings. 
There are no standard metrics to measure over-fitting or over-generalization in grammar generation, so we devise $4$ metrics based around the number of production rules used in parsing the positive examples. 

First, let $\Pi_\mathcal{P} \subseteq \Pi$ be the set of production rules that are used in the left-most derivations of all positive examples in $\mathcal{P}$. That is, the set of rules in $\Pi$ which occur in a sequence of rules $S \rightarrow \alpha_1 \rightarrow \ldots \rightarrow \alpha_n \rightarrow  p $ where $p \in \mathcal{P}$, and all rules expand the left-most non-terminal in $\alpha_1, \ldots, \alpha_n$. 

We report metrics across only \emph{solved} challenges, i.e., a challenge where $\guess$ is syntactically and semantically correct. Our four metrics are defined as follows\footnote{Refer to Appendix~\ref{appendix:grammar_quality_metrics} for the details of formal definitions.}:
\begin{squishitem}
    \item $\diff(C)$, calculates the difference between the number of production rules in $\guess$ used in parsing the positive examples and the number of production rules used by $\bnfref$ for a given challenge, i.e., $|\Pi^{ref}_\mathcal{P}| - |\Pi^*_\mathcal{P}|$. A grammar that uses substantially fewer production rules has probably overfitted to the examples, and a grammar that uses substantially more production rules may have over-generalized. We report the average of $\diff$ across all solved challenges as $\diff^\diamond$.
    \item $\of$ estimates over-fitting by counting the percentage of solved challenges on which $\guess$ uses fewer than half the number of production rules used by $\bnfref$, i.e., the number of times $|\Pi^{ref}_\mathcal{P}| - |\Pi^*_\mathcal{P}| > \frac{|\Pi^{ref}_\mathcal{P}|}{2}$. 
    \item $\og$ estimates over-generalization by counting the percentage of solved challenges on which $\guess$ uses more than half the number of production rules used by $\bnfref$, i.e.,  $|\Pi^{ref}_\mathcal{P}| - |\Pi^*_\mathcal{P}| < -\frac{|\Pi^{ref}_\mathcal{P}|}{2}$.
    \item $\tu$ calculates the proportion of production rules that are used in parsing the positive examples for a given challenge, i.e., $\frac{|\Pi^*_\mathcal{P}|}{\Pi^*}$, indicating the utility. We report the average of $\tu$ across all solved challenges as $TU^\diamond$.
\end{squishitem}

\subsection{Baselines}
To establish baselines, we adopted two approaches, Direct Prompting (DP) and Optimization of the BNF Parser for LLM-Friendly Feedback (OPF).

In DP, we directly prompted LLMs with positive and negative examples, asking them to produce grammars that accept all positives and reject all negatives\footnote{Refer to Appendix~\ref{appendix:DP} for the details of DP.}.

In OPF, inspired by Reflexion~\cite{shinn2023reflexion} and Self-Refine~\cite{madaan2023selfrefine}, we further optimized the BNF parser by enabling it to provide more LLM-friendly error messages as feedback, aiming to improve the performance of grammar generation for LLMs\footnote{Refer to Appendix~\ref{appendix:OPF} for the details of OPF.}. 

\subsection{Experiment Settings}
For a comprehensive evaluation, we selected a total of 8 LLMs, ensuring a diverse selection of both open- and closed-source LLMs, along with LLMs with varying parameter sizes and LLMs specifically designed for code generation. 

Specifically, we selected two closed-source LLMs, \textit{GPT-4o}~\cite{openai2024gpt4technicalreport} and \textit{GPT-3.5-Turbo}~\cite{brown2020languagemodelsfewshotlearners}. For the other 6 open-source LLMs, we note them in the notation: \text{\{\textit{model\_name}\}\textit{:}\{\textit{parameter\_size}\}\textit{-}\{\textit{model\_type}\}.} We selected the following LLMs: \textit{Llama3:70b-Instruct}~\cite{grattafiori2024llama3herdmodels}, \textit{Qwen:72b-Chat}~\cite{bai2023qwentechnicalreport}, \textit{Gemma2:27b-Instruct}~\cite{gemmateam2024gemma2improvingopen}, \textit{Mistral:7b-Instruct}~\cite{jiang2023mistral7b}, \textit{Codestral:22b}~\cite{mistral2024codestral}, and \textit{Starcoder2:15b-Instruct}~\cite{lozhkov2024starcoder2stackv2}. Among the selected LLMs, \textit{Starcoder2:15b-Instruct} and \textit{Codestral:22b} are LLMs for code generation.

For the DP baseline, we set the temperature to 0 and the maximum token to 2000.  

For the OPF baseline\footnote{Refer to Appendix~\ref{appendix:OPF} for the reason of setting the temperature greater than 0 for OPF.}, we set the temperature to 0.3, the maximum token to 2000, and the maximum number of turns, \textit{max\_turns}, to 5.

\subsection{Results \& Analysis}
The results of $\sx$ and $\se$ for the 8 LLMs are presented in Table~\ref{tab:results_for_all}. We observed that \textit{GPT-4o}, \textit{GPT-3.5-Turbo}, and \textit{Gemma2:27b-Instruct} achieve relatively high $\sx$, while the other LLMs generally fare worse in DP baseline. Compared to the DP baseline, applying OPF leads to a significant enhancement in $\sx$ for both \textit{Mistral:7b-Instruct} and \textit{Codestral:22b}, yielding an 18\% improvement for each of them, suggesting that parser feedback can help increase $\sx$. Nevertheless, for most LLMs, OPF yields only slight gains in $\sx$. Furthermore, it is worth noting that for \textit{Starcoder2:15b-Instruct}, $\sx$ decreases by 16\% after applying OPF. This indicates that feedback from the parser can sometimes lead to performance degradation if LLMs fail to interpret the feedback correctly. 

Although several LLMs attain high $\sx$, their $\se$ remains low in DP, with an average of 39\%. For example, \textit{GPT-3.5-Turbo} has 94\% $\sx$ but only 37\% $\se$. With OPF, LLMs like \textit{Mistral:7b-Instruct} and \textit{Codestral:22b} achieve notable $\se$ improvement with enhanced $\sx$, illustrating that improved $\sx$ can positively influence $\se$. However, for most LLMs, OPF yields only marginal $\se$ gains, particularly in the case that their $\sx$ is already high and OPF fails to contribute significant $\sx$ improvement.

Noticeably, although OPF helps improve $\sx$, a significant gap still persists between $\sx$ and $\se$ for most LLMs. For example, even after OPF, \textit{GPT-3.5-Turbo} maintains a 57\% gap between $\sx$ and $\se$, and \textit{Starcoder2:15b-Instruct} has a 40\% gap. Since the ultimate objective is to improve $\se$, a more advanced approach is needed to address this limitation.

Furthermore, as other four metrics shown in Table~\ref{tab:other_metrics}, on average, for most of LLMs, the lower $\diff^\diamond$, $\of$, and $\og$ indicate negligible over-fitting or over-generalization issues, and the higher $\tu^\diamond$ means LLMs are not predisposed to generate irrelevant production rules. Nevertheless, as the number of production rules increases, $\diff^\diamond$ and $\of$ exhibit slight increases, indicating a tendency toward mild over-fitting, while no significant over-generalization issues are observed~\footnote{Refer to Appendix~\ref{appendix:fgi_results} and~\ref{appendix:tu_results} for more details.}.

In addition to the quantitative metrics, we also conducted a qualitative analysis by examining the generated grammars. We compared the grammars containing syntactic errors generated in the DP but corrected after the OPF. We primarily found four significant issues causing lower $SX$: unsupported symbols such as injecting quantifiers like ``*'' and ``?'' or character classes like ``[a-z]'', erroneously introduced and misplaced brackets such as wrapping two terminals with round and square brackets, failing to wrap non-terminal with angle brackets, and omission the separators ``|'' between sentential forms. While the the first two issues are rampant across most LLMs, the third issue is mainly found in \textit{Mistral:7b-Instruct} and the last one is sporadic. For most LLMs, OPF can occasionally mitigate these issues, but insignificantly. Furthermore, we also observed LLMs show an ability to recognize keywords in the examples, treating them as complete terminals rather than decomposing them into multiple terminals as individual characters. For example, they treat ``if'' and ``SELECT'' as complete terminals rather than splitting them into multiple characters. This may be benefited from the common sense acquired by LLMs from the corpus~\cite{llm_few_shot}.

\begin{table*}
\centering
\begin{tabularx}{\textwidth}{l *{6}{>{\centering\arraybackslash}X}}
\toprule
{\textbf{Models}} & \multicolumn{3}{c}{\textbf{Syntax Correctness ($\sx$)}} & \multicolumn{3}{c}{\textbf{Semantics Correctness($\se$)}} \\
\cmidrule(lr){2-4} \cmidrule(lr){5-7}
 & DP & OPF & \textbf{\NAME} & DP & OPF & \textbf{\NAME} \\
\midrule
GPT-4o 
& 93 
& \textbf{97}~\textcolor{blue}{\scriptsize$\uparrow$4} 
& 96~\textcolor{blue}{\scriptsize$\uparrow$3}~\textcolor{red}{\scriptsize$~\downarrow$1} 
& 84 
& 85~\textcolor{blue}{\scriptsize$\uparrow$1}
& \textbf{93}~\textcolor{blue}{\scriptsize$\uparrow$9}~\textcolor{red}{\scriptsize$~\uparrow$8}  \\
GPT-3.5-Turbo 
& 94 
& 95~\textcolor{blue}{\scriptsize$\uparrow$1} 
& \textbf{99}~\textcolor{blue}{\scriptsize$\uparrow$5}~\textcolor{red}{\scriptsize$~\uparrow$4} 
& 37 
& 38~\textcolor{blue}{\scriptsize$\uparrow$1}
& \textbf{61}~\textcolor{blue}{\scriptsize$\uparrow$24}~\textcolor{red}{\scriptsize$\uparrow$23}  \\
Llama3:70b-Instruct 
& 57 
& 61~\textcolor{blue}{\scriptsize$\uparrow$4}
& \textbf{75}~\textcolor{blue}{\scriptsize$\uparrow$18}~\textcolor{red}{\scriptsize$\uparrow$14} 
& 41 
& 42~\textcolor{blue}{\scriptsize$\uparrow$1} 
& \textbf{61}~\textcolor{blue}{\scriptsize$\uparrow$20}~\textcolor{red}{\scriptsize$\uparrow$19}  \\
Qwen:72b-Chat 
& 47 
& 49~\textcolor{blue}{\scriptsize$\uparrow$2}
& \textbf{76}~\textcolor{blue}{\scriptsize$\uparrow$29}~\textcolor{red}{\scriptsize$\uparrow$27}
& 20 
& 21~\textcolor{blue}{\scriptsize$\uparrow$1}
& \textbf{38}~\textcolor{blue}{\scriptsize$\uparrow$18}~\textcolor{red}{\scriptsize$\uparrow$17} \\
Mistral:7b-Instruct
& \phantom{0}1 
& \textbf{19}~\textcolor{blue}{\scriptsize$\uparrow$18} 
& \phantom{0}1~\textcolor{blue}{\scriptsize$-$}~\textcolor{red}{\scriptsize$\downarrow$18}
& \phantom{0}0 
& \phantom{0}\textbf{8}~\textcolor{blue}{\scriptsize$\uparrow$8}
& \phantom{0}1~\textcolor{blue}{\scriptsize$\uparrow$1}~\textcolor{red}{\scriptsize$\downarrow$7}\\
Gemma2:27b-Instruct 
& 91 
& 92~\textcolor{blue}{\scriptsize$\uparrow$1} 
& \textbf{98}~\textcolor{blue}{\scriptsize$\uparrow$7}~\textcolor{red}{\scriptsize$\uparrow$6}
& 56 
& 57~\textcolor{blue}{\scriptsize$\uparrow$1} 
& \textbf{79}~\textcolor{blue}{\scriptsize$\uparrow$23}~\textcolor{red}{\scriptsize$\uparrow$22} \\
Starcoder2:15b-Instruct 
& 76 
& 60~\textcolor{blue}{\scriptsize$\downarrow$16} 
& \textbf{98}~\textcolor{blue}{\scriptsize$\uparrow$22}~\textcolor{red}{\scriptsize$\uparrow$38}
& 30 
& 20~\textcolor{blue}{\scriptsize$\downarrow$10} 
& \textbf{44}~\textcolor{blue}{\scriptsize$\uparrow$14}~\textcolor{red}{\scriptsize$\uparrow$24} \\
Codestral:22b 
& 53 
& 71~\textcolor{blue}{\scriptsize$\uparrow$18} 
& \textbf{80}~\textcolor{blue}{\scriptsize$\uparrow$27}~\textcolor{red}{\scriptsize$\uparrow$9}
& 44 
& 52~\textcolor{blue}{\scriptsize$\uparrow$8} 
& \textbf{67}~\textcolor{blue}{\scriptsize$\uparrow$23}~\textcolor{red}{\scriptsize$\uparrow$15} \\
\bottomrule
\end{tabularx}
\caption{The results of syntax and semantic correctness for LLMs grammar generation are presented as percentages (\%). For each LLM, the best syntax and semantic correctness are highlighted with bold font. Blue arrows \textcolor{blue}{$\uparrow\downarrow$} represent performance differences relative to the DP baseline, while red arrows \textcolor{red}{$\uparrow\downarrow$} indicate differences relative to the OPF baseline.}
\label{tab:results_for_all}
\end{table*}

% LLM-Driven Heuristic Mutation Genetic Algorithm
\section{LLM-Driven Hybrid Genetic Algorithm}
\label{sec:ldhm_ga}
The results of the baselines demonstrate the unsatisfactory performance of LLMs in grammar generation. To address this, we propose an LLM-driven hybrid genetic algorithm, namely \NAME, a novel algorithm inspired by the concept of genetic algorithms. The following sections detail \NAME and elaborate on our experiment settings, results, and analysis.

\subsection{Methodology}
\NAME consists of four main components: Fitness, Selection, Crossover, and Mutation. It begins by prompting an LLM to generate an initial population of candidate grammars from positive and negative examples. In each generation, the $\mathrm{Fitness}$ function scores each candidate and the $\mathrm{Select}$ function chooses a subset of the population. To form a new population, the $\mathrm{Cross}$ function operates on two randomly selected candidates from this subset to generate a new candidate, which is then modified by the $\mathrm{Mutate}$ function and added to the new population, until the maximum population size is reached. The new population then advances to the next generation. We provide pseudocode in Algorithm~\ref{algo:ldhga} in Appendix~\ref{appendix:LDHGA}. We detail each component as follows:

\paragraph{Fitness}
Given a generated grammar $\guess$, if it is syntactically incorrect, it is assigned a score of $-1$. 
% Thus, the fitness function for an empty $\mathcal{E}^{\mathcal{G}}$ is given as $\mathrm{Fitness}(\mathcal{G},\mathcal{P},\mathcal{N}) = -1$.
For a valid grammar, we define two indicator functions:
\begin{align*}
\mathbb{I}_{\mathcal{A}}(\guess, p) &=
\begin{cases}
1 & \text{if } p \in \languageOf{\guess},\\
0 & \text{otherwise}.
\end{cases}\\
\mathbb{I}_{\mathcal{R}}(\guess, n) &=
\begin{cases}
1 & \text{if } n \notin \languageOf{\guess},\\
0 & \text{otherwise}.
\end{cases}
\end{align*}
The $\mathrm{Fitness}$ function $\mathrm{Fitness}(\guess,\mathcal{P},\mathcal{N})$ is then defined as: 
\[
\sum_{p_i\in \mathcal{P}} 
\mathbb{I}_{\mathcal{A}}(\guess,\,p_i) 
+ \sum_{n_i \in \mathcal{N}} 
\mathbb{I}_{\mathcal{R}}(\guess,\,n_i). 
\]
% where 
% $\mathcal{A}=\mathbb{I}_{\mathcal{A}}(\mathcal{E}^{\mathcal{G}},\,p_i)$ 
% and $\mathcal{D}=\mathbb{I}_{\mathcal{R}}(\mathcal{E}^\mathcal{G},\,n_i)$.

\paragraph{Selection}
 Let $\grammarset = \{\guess_1, \guess_2, \ldots, \guess_k\}$ be a population of candidate grammars.
% 
 % $E = [\mathcal{E}^{\mathcal{G}_1},\mathcal{E}^{\mathcal{G}_2},\dots,\mathcal{E}^{\mathcal{G}_k}]$ 
 % be a set of grammar edge sets representing a population of candidates. 
 %
 Each $\guess_i \in \grammarset$ is assigned a fitness score $s_i \in S$ by the $\mathrm{Fitness}$ function, where $S = [s_1,s_2,\dots,s_k]$ is a sequence of their corresponding scores. We define the $\mathrm{Select}$ function as:
 $$
\mathrm{Select}(\grammarset,S) = \{\guess_{\sigma(1)}, \guess_{\sigma(2)}, \dots, \guess_{\sigma(\frac{k}{2})}\},
$$
where $\sigma : \{1, 2, \dots, k\} \to \{1, 2, \dots, k\}$ is a permutation of the indices such that: $s_{\sigma(1)} \geq s_{\sigma(2)} \geq \dots \geq s_{\sigma(k)}$.
% $$
% \mathrm{Select}(E,S) = [\mathcal{E}^{\mathcal{G}_{\sigma(1)}}, \mathcal{E}^{\mathcal{G}_{\sigma(2)}}, \dots, \mathcal{E}^{\mathcal{G}_{\sigma(\lfloor k/2 \rfloor)}}],
% $$
% where $\sigma : \{1, 2, \dots, k\} \to \{1, 2, \dots, k\}$ is a permutation of the indices such that: \[s_{\sigma(1)} \geq s_{\sigma(2)} \geq \dots \geq s_{\sigma(k)}\]
%
Hence, half of the candidates from the population are selected in decreasing order of their fitness scores.

\paragraph{Crossover}
The crossover function splices the production rules from two grammars together, at a randomly chosen splicing point. 
Let 
$\guess_a$ and $\guess_b$ be two candidate grammars, with rules $R^*_a$ and $R^*_b$ respectively.
If both $R$ are empty, we return one grammar at random. If one grammar has a non-empty $R$, we return that grammar. Otherwise, we apply the crossover function with probability $\rho$ (the crossover rate).

The crossover operation works in the following way. Let $\ell = \min(|R^*_a|, |R^*_b|)$.
First sample a crossover point
$
w \sim \mathrm{Uniform}(\{1,2,\dots,\ell\}).
$
Let $R_a^* = r^a_1, r^a_2 \ldots $ be the rules from $\guess_a$ and $R_b^*$ be the rules from $\guess_b$. We take the first $w-1$ rule sets from $R_a^*$ and then prefix these to the last $n-w$ rule sets from $R_b^*$, where $n = |R^*_b|$. This generates a new rule set $R' = \{r^1_a, \ldots, r^{w-1}_a, r^w_b, \ldots r^n_b\}$. 

Crossover returns a new grammar $\bnf' = (V', \Sigma', \Pi', S_a, R')$, where $V'$ and $\Sigma'$ are all nonterminal and terminal symbols in $R'$, $\Pi'$ is all rules in $R'$, and $S_a$ is the start symbol from $\guess_a$.

\paragraph{Mutation}
We use two mutation methods: mutation by LLM, and local mutation. 
We chose whether to mutate a grammar at all, with probability $\mu$ (the mutation rate). If the grammar has no production rules, we apply the LLM mutation. Otherwise, we apply local mutation with a probability $0.5$ and LLM mutation otherwise.

\paragraph{LLM-Driven Mutation}
The LLM-driven mutation 
uses $\guess$, $\mathcal{P}$, and $\mathcal{N}$ to prompt\footnote{Refer to Prompt Template~\ref{prompt:llm_driven_mutation} in Appendix~\ref{appendix:LDHGA} for the prompt we designed for LLM-driven mutation.} an LLM to produce a new grammar. In this approach, we expect that, with the knowledge and experience obtained by training on vast corpora, LLMs can heuristically provide more novel and dramatic modifications such as introducing new terminals or non-terminals, adding or removing production rules, or reshaping the structure of grammars, which is hard to approach with local mutation.

\paragraph{Local Mutation}
The local mutation is designed to produce incremental, targeted alterations to the grammar while preserving the majority of its original form. It is less flexible than LLM-driven mutation and it is unable to introduce new non-terminals or to drastically restructure grammars. However, it can not only potentially find a grammar candidate but also provide insights for LLM-driven mutation. Given a grammar $\guess$, with a set of sets of production rules $R^*$, local mutation comprises the following steps:

\begin{itemize}
    % \item \textbf{Rule Extraction.} 
    % Let $\mathcal{E}(\mathcal{G}) = [(x_1, Y_1), \dots, (x_l, Y_l)]$ denote a sequence of $l$ elements in $\mathcal{G}$, where $(v_i, A_i) \in \mathrm{Comps}(\mathcal{G})$ is a element in $\mathcal{G}$, $v_i$ is a non-terminal symbol on the left side of a element, and $A_i$ is the corresponding sequence of alternatives on the right side of a element.

    \item \textbf{Rule set selection} 
    We sample an integer $i \;\sim\; \mathrm{Uniform}(\{1,\dots,|R|\})$. 
    This index $i$ chooses the rule set  $r_i \in R^*$ that will be mutated.

    \item \textbf{Shuffle} 
    % We designed two types of alterations, $\mathrm{Shuffle}$ and $\mathrm{SpaceInsert}$. 
%
     The $\mathrm{Shuffle}$ mutation shuffles the order of symbols on the right-hand side of a production rule. 
     That is, each rule is a mapping from a non-terminal $v$ to a sequence of non-terminal and terminal symbols $(V \cup \Sigma)^*$, and we shuffle this sequence randomly. For example, the rule set:
    \verb+<e> ::= <e> "*" <e> | <e> "/" <e>+ with infix operators, may be shuffled to 
    \verb+<e> ::= "*" <e> <e> | "/" <e> <e>+, switching to prefix operators.
    
    Shuffle is applied to all production rules in $r_i$.

    % For $\mathrm{Shuffle}$, each rule  $y \in Y_r$ is shuffled in place (i.e., its internal sequence of symbols is permuted randomly). Formally,
    % \begin{align*}
    %     \mathrm{Shuffle}(Y_r) \;=\; \bigl[s_{\pi(1)},\, s_{\pi(2)},\, \dots,\, s_{\pi(|Y_r|)}\bigr],
    % \end{align*}
    % where $\pi$ is a permutation on the set $\{1,2,\dots,|Y_r|\}$ chosen uniformly at random. After $\mathrm{Shuffle}$, the production rules of $\mathcal{G}$ are changed leading to modifying the original $\mathcal{G}=(V,\Sigma,R,S)$ to $\mathcal{G}'=(V,\Sigma,R',S)$.

    \item \textbf{Space Insertion}
    The $\mathrm{SpaceInsert}$ mutation inserts a randomly chosen number of whitespace terminal symbols \verb+"␣"+ into the right-hand side of a production rule. Given a rule $v_i \rightarrow \alpha$, Shuffle chooses the number of whitespace terminals to be inserted by sampling $I \sim \textrm{Uniform}(0,|\alpha|)$, where $|\alpha|$ is the number of symbols in $\alpha$. Each space is inserted before or after a randomly chosen symbol in $\alpha$.
    As an example, the rule set:
    \verb+<s> ::= <noun> <verb>+ may be changed to 
    \verb+<s> ::= <noun> "␣" <verb>+.

    For each production rule in $r_i$, we randomly decide whether $\mathrm{SpaceInsert}$ should be applied to that rule with a low probability\footnote{We fix it to 0.1.}.

\end{itemize}
    The $\mathrm{Shuffle}$ alteration heuristic is motivated by two key insights. First, shuffling the right-hand side of production rules may yield a grammar that accepts more positive examples and rejects more negative ones.  Second, although $\mathrm{Shuffle}$ may rarely yield a better candidate for a complex target grammar, the new variant of the grammar produced from $\mathrm{Shuffle}$ is expected to provide alternative perspectives and new insights for LLMs to help generate subsequent grammars in future generations.
    
    The $\mathrm{SpaceInsert}$ alteration was introduced because some LLMs tended to omit explicit space symbols between symbols in an alternative, resulting in degraded grammar generation, even if they were prompted to pay attention to space inclusion\footnote{Refer to Prompt Template~\ref{prompt:direct_prompting} in Appendix~\ref{appendix:DP} for details.}. We expect that incorporating $\mathrm{SpaceInsert}$ will offer insights for LLMs of the explicit inclusion of spaces to thereby enhance performance.
    
    % For instance, when tasked with generating a simple grammar for a calculator that accepts only Polish notation, some LLMs produce a grammar such as:
    % $$
    % \texttt{<cal>}\vcentcolon \vcentcolon=\texttt{<number>} \texttt{ "+" } \texttt{<number>}
    % $$
    % However, after $\mathrm{Shuffle}$, a new grammar can be obtained:
    % $$
    % \texttt{<cal>}\vcentcolon \vcentcolon=\texttt{"+" }\texttt{<number> }\texttt{<number>}
    % $$ 

\subsection{Experiment Settings}
In addition to a set of positive and negative examples and an LLM, \NAME takes four parameters: \emph{population size} (grammars per generation), \emph{generations} (number of evolution iterations), \emph{crossover rate} (probability of crossover), \emph{mutation rate} (probability of mutation). In our experiments, we set these to 10, 5, 0.7, and 0.3, respectively.

We selected the same 8 LLMs as the DP and OPF baselines, setting maximum tokens to 2000 and temperature to 0.7. In \NAME, a nonzero temperature is necessary for diversity. 
% Nevertheless, we expect that after multiple generations, it can find an optimal grammar. 
To ensure it does not significantly impact results and show its robustness, we further repeat the experiments 5 times with \textit{GPT-3.5-Turbo} and \textit{GPT-4o}.

\subsection{Results \& Analysis}

% Diff, OF, and OG
\begin{table}[t]
  \centering
  \begin{tabular}{lcccc}
    \toprule
    \textbf{Methods} & $\diff^{\diamond}$ & $\of$ & $\og$ & $\tu^\diamond$\\
    \midrule
     DP     & 1.12 &  3.83 & 0.63 &  88.74 \\
     OPF    & 1.10 &  4.72 & 1.31  & 90.76 \\
     \NAME  & 1.19 &  4.44  & 0.92  & 91.27  \\
    \bottomrule
  \end{tabular}
  \caption{The averages of $\diff^\diamond$, $\of$(\%), $\og$(\%), and $\tu^\diamond$(\%) across all LLMs.}
  \label{tab:other_metrics}
\end{table}

As shown in Table~\ref{tab:results_for_all}, \NAME substantially boosts $\sx$ for most LLMs. For example, \textit{Qwen:72b-Instruct} gains 29\% over DP and 27\% over OPF. Even for LLMs already improved by OPF, such as \textit{Codestral:22b} with an 18\% improvement, \NAME adds an additional 9\%. Meanwhile, \textit{Starcoder2:15b-Instruct}, which experiences a 16\% drop under OPF, achieves a 22\% improvement compared to DP and a 38\% improvement over OPF, with \NAME. On average, it improves $\sx$ 13.88\% compared to DP and 9.88\% over OPF.

While enhancing $\sx$ is essential, the ultimate objective is to improve $\se$. As shown in Table~\ref{tab:results_for_all}, our method significantly boosts $\se$ for all LLMs except \textit{Mistral:7b-Instruct}. For example, with \NAME, \textit{GPT-4o} rises from 84\% with DP and 85\% with OPF to 93\%, and \textit{GPT-3.5-Turbo}, noted for low semantic accuracy, increases by 24\% over DP and 23\% over OPF. Notably, although the contribution from the enhancement of $\sx$ is essential to $\se$, \NAME does not rely solely on enhancing $\sx$ to achieve significant improvement of $\se$, as five LLMs demonstrated higher $\se$ increases than their $\sx$. Across the selected LLMs, the average $\se$ improvement is 16.5\% compared to DP and 15.13\% compared to OPF. 

We further analyzed the performance as the number of non-terminals and production rules increases. For non-terminals, we partition the dataset into 3 groups: $C_1$ (1–3 non-terminals), $C_2$ (4-6 non-terminals), and $C_3$ (7-9 non-terminals). For production rules, we split the dataset into another 3 groups: $P_1$ (1–6 production rules), $P_2$ (6-15 production rules), and $P_3$ (greater than 16 production rules). We observed that the performance of LLMs decreases as the number of non-terminals and production rules increases. Nevertheless, \NAME still substantially improves both $\sx$ and $\se$\footnote{Refer to Tables~\ref{tab:nonterminals_results} and~\ref{tab:prs_results} in Appendix~\ref{appendix:nonterminals_results} and~\ref{appendix:prs_results} for details.}.

Furthermore, as shown in Table~\ref{tab:other_metrics}, $\of$ does not significantly increase in \NAME, indicating that the substantial improvements observed in both the $\sx$ and $\se$ for \NAME are not attributed to overfitting\footnote{Refer to Tables~\ref{tab:fgi_diff},~\ref{tab:fgi_of}, and~\ref{tab:fgi_og} in Appendix~\ref{appendix:fgi_results} for the more details.}.

Moreover, we also conducted a qualitative analysis of how \NAME improves $SX$ and $SE$. For $SX$, \NAME significantly reduces the issues of unsupported symbols injection and misplaced brackets. However, it fails to address the issue of unwrapped non-terminals, which is also the issue mainly happened in \textit{Mistral:7b-Instruct}. Unlike OPF, which benefits from more explicit syntax error feedback, our approach lacks such direct syntax corrective guidance, meaning that if an LLM inherently struggles to generate syntactically correct grammars, our method may fail to produce valid candidates and process evolution, thereby lowering both $SX$ and $SE$. Nevertheless, as long as at least a few candidates are generated in correct syntax, \NAME can optimize their generations during the evolutionary process to mitigate the aforementioned issues and improve $\sx$. For $\se$, attributing the significant improvement is complex. However, we still observed two phenomena. First, after applying \NAME, terminals that were not previously considered in the grammars generated by DP or OPF have been introduced. Second, the semantic errors that were caused by the absence of space terminals in DP or OPF have been alleviated.

Due to the relatively high temperature of 0.7 used for \NAME, we repeated the experiments 5 times with \textit{GPT-4o} and \textit{GPT-3.5-Turbo}\footnote{Refer to Table~\ref{tab:5_times} in Appendix~\ref{appendix:multiple_experiments} for details.} to ensure robustness. The averages of $\sx$ for \textit{GPT-4o} and \textit{GPT-3.5-Turbo} are 95.8\% and 98.6\%, with standard deviations 0.4\% and 0.49\% respectively, while the averages of $\se$ are 93.2\% and 61.6\% with standard deviations 0.4\% and 0.49\% respectively. These results indicate that setting the temperature to 0.7 has a negligible impact on performance and show the robustness of \NAME.

% Conclusion
\section{Conclusion}
To explore the few-shot grammar generation ability of LLMs, we constructed a dedicated dataset consisting of 540 challenges, devised and adopted 6 metrics, and evaluated 8 various LLMs. Due to their unsatisfactory performance, we introduced \NAME, an LLM-driven hybrid genetic algorithm for grammar generation. Our results indicate that \NAME significantly enhances both syntax and semantic correctness compared to the two baselines. We believe this work provides valuable insights into LLM-based grammar generation and highlights the potential of LLM-driven hybrid genetic algorithms in this domain.

% Limitations
\section{Limitations}
\label{sec:limitations}
We discuss several limitations and concerns in this work, revealing potential challenges, constraints, and confusion. 

% First, each challenge in the proposed dataset only consists of 3 positive and 3 negative examples to emphasizing grammar generation in few-shot manner. Typically, other grammar inference algorithms require characteristic samples to uniquely determine the target grammar~\cite{de2010grammatical}. In contrast, we allow general samples that may not uniquely represent a single grammar. Consequently, a given set of examples can correspond to multiple grammars that accept all positives and reject all negatives. This ambiguity motivated the introduction of the $FGI$ metric, which measures how closely a generated grammar approximates the reference grammar. Future work could explore or construct datasets with more examples per challenge or even incorporate characteristic samples to further evaluate LLM performance in grammar generation.

First, although the results indicate that \textit{GPT-4o} exhibits remarkable $SX$ and $SE$, it is important to note that these results may be attributable to the use of GPT-4o during dataset construction. Nonetheless, even though \textit{GPT-4o} already demonstrates excellent performance, \NAME can still enhance it significantly.

Second, as demonstrated, our method does not outperform OPF for \textit{Mistral:7b-Instruct} in $SX$ and $SE$ due to its inherent failure to generate syntactically correct grammars. Nevertheless, our approach yields significant $SX$ and $SE$ improvements for all other LLMs. We also propose to combine syntactical feedback and \NAME to mitigate this limitation and further improve the performance.

Third, one may argue that given any finite set of positive and negative examples, it could always be possible to construct a regular grammar rather than a CFG that can accept all positives and reject all negatives. However, such an approach may function more like a classifier rather than a grammar and may lack applicability in subsequent tasks, such as constructing an abstract syntax tree. 

Finally, in this work, we primarily focus on LLM-based few-shot grammar generation without comparing algorithms that are not LLM-based. The reasons behind this are that most algorithms require a large set of characteristic examples to uniquely determine the target grammar~\cite{de2010grammatical}. Instead, we do not impose such constraints on our example set and hypothesize that the experience and knowledge acquired from corpus can enable LLMs to handle few-shot grammar generation tasks. Consequently, those algorithms may not be directly applicable. In addition, since we focus on the exploration and improvement of the ability of LLMs in few-shot grammar generation, we construct two LLM-based baselines for fair comparison.

\bibliography{anthology,custom}

\begin{thebibliography}{45}
\providecommand{\natexlab}[1]{#1}

\bibitem[{Aho(2007)}]{Aho_Aho_2007}
 2007.
\newblock \emph{Compilers: principles, techniques, tools}, 2nd ed edition.
\newblock Pearson/Addison Wesley, Boston.

\bibitem[{Backus(1959)}]{backus1959syntax}
John~W Backus. 1959.
\newblock The syntax and the semantics of the proposed international algebraic language of the zurich acm-gamm conference.
\newblock In \emph{ICIP Proceedings}, pages 125--132.

\bibitem[{Backus et~al.(1960)Backus, Bauer, Green, Katz, McCarthy, Perlis, Rutishauser, Samelson, Vauquois, Wegstein et~al.}]{backus1960report}
John~W Backus, Friedrich~L Bauer, Julien Green, Charles Katz, John McCarthy, Alan~J Perlis, Heinz Rutishauser, Klaus Samelson, Bernard Vauquois, Joseph~Henry Wegstein, et~al. 1960.
\newblock Report on the algorithmic language algol 60.
\newblock \emph{Communications of the ACM}, 3(5):299--311.

\bibitem[{Bai et~al.(2023)Bai, Bai, Chu, Cui, Dang, Deng, Fan, Ge, Han, Huang, Hui, Ji, Li, Lin, Lin, Liu, Liu, Lu, Lu, Ma, Men, Ren, Ren, Tan, Tan, Tu, Wang, Wang, Wang, Wu, Xu, Xu, Yang, Yang, Yang, Yang, Yao, Yu, Yuan, Yuan, Zhang, Zhang, Zhang, Zhang, Zhou, Zhou, Zhou, and Zhu}]{bai2023qwentechnicalreport}
Jinze Bai, Shuai Bai, Yunfei Chu, Zeyu Cui, Kai Dang, Xiaodong Deng, Yang Fan, Wenbin Ge, Yu~Han, Fei Huang, Binyuan Hui, Luo Ji, Mei Li, Junyang Lin, Runji Lin, Dayiheng Liu, Gao Liu, Chengqiang Lu, Keming Lu, Jianxin Ma, Rui Men, Xingzhang Ren, Xuancheng Ren, Chuanqi Tan, Sinan Tan, Jianhong Tu, Peng Wang, Shijie Wang, Wei Wang, Shengguang Wu, Benfeng Xu, Jin Xu, An~Yang, Hao Yang, Jian Yang, Shusheng Yang, Yang Yao, Bowen Yu, Hongyi Yuan, Zheng Yuan, Jianwei Zhang, Xingxuan Zhang, Yichang Zhang, Zhenru Zhang, Chang Zhou, Jingren Zhou, Xiaohuan Zhou, and Tianhang Zhu. 2023.
\newblock \href {https://arxiv.org/abs/2309.16609} {Qwen technical report}.
\newblock \emph{Preprint}, arXiv:2309.16609.

\bibitem[{Beurer-Kellner et~al.(2024)Beurer-Kellner, Fischer, and Vechev}]{BeurerKellner2024GuidingLT}
Luca Beurer-Kellner, Marc Fischer, and Martin~T. Vechev. 2024.
\newblock \href {https://api.semanticscholar.org/CorpusID:268363645} {Guiding llms the right way: Fast, non-invasive constrained generation}.
\newblock \emph{ArXiv}, abs/2403.06988.

\bibitem[{Brown et~al.(2020{\natexlab{a}})Brown, Mann, Ryder, Subbiah, Kaplan, Dhariwal, Neelakantan, Shyam, Sastry, Askell, Agarwal, Herbert-Voss, Krueger, Henighan, Child, Ramesh, Ziegler, Wu, Winter, Hesse, Chen, Sigler, Litwin, Gray, Chess, Clark, Berner, McCandlish, Radford, Sutskever, and Amodei}]{brown2020languagemodelsfewshotlearners}
Tom~B. Brown, Benjamin Mann, Nick Ryder, Melanie Subbiah, Jared Kaplan, Prafulla Dhariwal, Arvind Neelakantan, Pranav Shyam, Girish Sastry, Amanda Askell, Sandhini Agarwal, Ariel Herbert-Voss, Gretchen Krueger, Tom Henighan, Rewon Child, Aditya Ramesh, Daniel~M. Ziegler, Jeffrey Wu, Clemens Winter, Christopher Hesse, Mark Chen, Eric Sigler, Mateusz Litwin, Scott Gray, Benjamin Chess, Jack Clark, Christopher Berner, Sam McCandlish, Alec Radford, Ilya Sutskever, and Dario Amodei. 2020{\natexlab{a}}.
\newblock \href {https://arxiv.org/abs/2005.14165} {Language models are few-shot learners}.
\newblock \emph{Preprint}, arXiv:2005.14165.

\bibitem[{Brown et~al.(2020{\natexlab{b}})Brown, Mann, Ryder, Subbiah, Kaplan, Dhariwal, Neelakantan, Shyam, Sastry, Askell, Agarwal, Herbert-Voss, Krueger, Henighan, Child, Ramesh, Ziegler, Wu, Winter, Hesse, Chen, Sigler, Litwin, Gray, Chess, Clark, Berner, McCandlish, Radford, Sutskever, and Amodei}]{llm_few_shot}
Tom~B. Brown, Benjamin Mann, Nick Ryder, Melanie Subbiah, Jared Kaplan, Prafulla Dhariwal, Arvind Neelakantan, Pranav Shyam, Girish Sastry, Amanda Askell, Sandhini Agarwal, Ariel Herbert-Voss, Gretchen Krueger, Tom Henighan, Rewon Child, Aditya Ramesh, Daniel~M. Ziegler, Jeffrey Wu, Clemens Winter, Christopher Hesse, Mark Chen, Eric Sigler, Mateusz Litwin, Scott Gray, Benjamin Chess, Jack Clark, Christopher Berner, Sam McCandlish, Alec Radford, Ilya Sutskever, and Dario Amodei. 2020{\natexlab{b}}.
\newblock Language models are few-shot learners.
\newblock In \emph{Proceedings of the 34th International Conference on Neural Information Processing Systems}, NIPS '20, Red Hook, NY, USA. Curran Associates Inc.

\bibitem[{Chen(1995)}]{chen1995bayesian}
Stanley~F Chen. 1995.
\newblock Bayesian grammar induction for language modeling.
\newblock \emph{arXiv preprint cmp-lg/9504034}.

\bibitem[{Chen et~al.(2023)Chen, Lin, Schärli, and Zhou}]{Chen2023Teaching}
Xinyun Chen, Maxwell Lin, Nathanael Schärli, and Denny Zhou. 2023.
\newblock \href {https://doi.org/10.48550/arXiv.2304.05128} {Teaching large language models to self-debug}.
\newblock \emph{ArXiv}, abs/2304.05128.

\bibitem[{Chomsky(1956)}]{Chomsky_1956}
N.~Chomsky. 1956.
\newblock \href {https://doi.org/10.1109/TIT.1956.1056813} {Three models for the description of language}.
\newblock \emph{IRE Transactions on Information Theory}, 2(3):113--124.

\bibitem[{Cohen et~al.(2017)Cohen, Caciularu, Rejwan, and Berant}]{Cohen2017InducingRG}
Mor Cohen, Avi Caciularu, Idan Rejwan, and Jonathan Berant. 2017.
\newblock \href {https://api.semanticscholar.org/CorpusID:8704099} {Inducing regular grammars using recurrent neural networks}.
\newblock \emph{ArXiv}, abs/1710.10453.

\bibitem[{De~la Higuera(2010)}]{de2010grammatical}
Colin De~la Higuera. 2010.
\newblock \emph{Grammatical inference: learning automata and grammars}.
\newblock Cambridge University Press.

\bibitem[{Dehaerne et~al.(2022)Dehaerne, Dey, Halder, De~Gendt, and Meert}]{dehaerne2022}
Enrique Dehaerne, Bappaditya Dey, Sandip Halder, Stefan De~Gendt, and Wannes Meert. 2022.
\newblock \href {https://doi.org/10.1109/ACCESS.2022.3196347} {Code generation using machine learning: A systematic review}.
\newblock \emph{IEEE Access}, 10:82434--82455.

\bibitem[{D’ulizia et~al.(2011)D’ulizia, Ferri, and Grifoni}]{Dulizia2011A}
Arianna D’ulizia, F.~Ferri, and P.~Grifoni. 2011.
\newblock \href {https://doi.org/10.1007/s10462-010-9199-1} {A survey of grammatical inference methods for natural language learning}.
\newblock \emph{Artificial Intelligence Review}, 36:1--27.

\bibitem[{D’Ulizia et~al.(2011)D’Ulizia, Ferri, and Grifoni}]{D_Ulizia_Ferri_Grifoni_2011}
Arianna D’Ulizia, Fernando Ferri, and Patrizia Grifoni. 2011.
\newblock \href {https://doi.org/10.1007/s10462-010-9199-1} {A survey of grammatical inference methods for natural language learning}.
\newblock \emph{Artificial Intelligence Review}, 36(1):1–27.

\bibitem[{Gaur and Saunshi(2023)}]{gaur-saunshi-2023-reasoning}
Vedant Gaur and Nikunj Saunshi. 2023.
\newblock \href {https://doi.org/10.18653/v1/2023.findings-acl.364} {Reasoning in large language models through symbolic math word problems}.
\newblock In \emph{Findings of the Association for Computational Linguistics: ACL 2023}, pages 5889--5903, Toronto, Canada. Association for Computational Linguistics.

\bibitem[{Grattafiori et~al.(2024)Grattafiori, Dubey, Jauhri, Pandey, Kadian, Al-Dahle, Letman, Mathur, Schelten, Vaughan, Yang, Fan, Goyal, Hartshorn, Yang, Mitra, Sravankumar, Korenev, Hinsvark, Rao, Zhang, Rodriguez, Gregerson, Spataru, Roziere, Biron, Tang, Chern, Caucheteux, Nayak, Bi, Marra, McConnell, Keller, Touret, Wu, Wong, Ferrer, Nikolaidis, Allonsius, Song, Pintz, Livshits, Wyatt, Esiobu, Choudhary, Mahajan, Garcia-Olano, Perino, Hupkes, Lakomkin, AlBadawy, Lobanova, Dinan, Smith, Radenovic, Guzmán, Zhang, Synnaeve, Lee, Anderson, Thattai, Nail, Mialon, Pang, Cucurell, Nguyen, Korevaar, Xu, Touvron, Zarov, Ibarra, Kloumann, Misra, Evtimov, Zhang, Copet, Lee, Geffert, Vranes, Park, Mahadeokar, Shah, van~der Linde, Billock, Hong, Lee, Fu, Chi, Huang, Liu, Wang, Yu, Bitton, Spisak, Park, Rocca, Johnstun, Saxe, Jia, Alwala, Prasad, Upasani, Plawiak, Li, Heafield, Stone, El-Arini, Iyer, Malik, Chiu, Bhalla, Lakhotia, Rantala-Yeary, van~der Maaten, Chen, Tan, Jenkins, Martin, Madaan, Malo, Blecher,
  Landzaat, de~Oliveira, Muzzi, Pasupuleti, Singh, Paluri, Kardas, Tsimpoukelli, Oldham, Rita, Pavlova, Kambadur, Lewis, Si, Singh, Hassan, Goyal, Torabi, Bashlykov, Bogoychev, Chatterji, Zhang, Duchenne, Çelebi, Alrassy, Zhang, Li, Vasic, Weng, Bhargava, Dubal, Krishnan, Koura, Xu, He, Dong, Srinivasan, Ganapathy, Calderer, Cabral, Stojnic, Raileanu, Maheswari, Girdhar, Patel, Sauvestre, Polidoro, Sumbaly, Taylor, Silva, Hou, Wang, Hosseini, Chennabasappa, Singh, Bell, Kim, Edunov, Nie, Narang, Raparthy, Shen, Wan, Bhosale, Zhang, Vandenhende, Batra, Whitman, Sootla, Collot, Gururangan, Borodinsky, Herman, Fowler, Sheasha, Georgiou, Scialom, Speckbacher, Mihaylov, Xiao, Karn, Goswami, Gupta, Ramanathan, Kerkez, Gonguet, Do, Vogeti, Albiero, Petrovic, Chu, Xiong, Fu, Meers, Martinet, Wang, Wang, Tan, Xia, Xie, Jia, Wang, Goldschlag, Gaur, Babaei, Wen, Song, Zhang, Li, Mao, Coudert, Yan, Chen, Papakipos, Singh, Srivastava, Jain, Kelsey, Shajnfeld, Gangidi, Victoria, Goldstand, Menon, Sharma, Boesenberg,
  Baevski, Feinstein, Kallet, Sangani, Teo, Yunus, Lupu, Alvarado, Caples, Gu, Ho, Poulton, Ryan, Ramchandani, Dong, Franco, Goyal, Saraf, Chowdhury, Gabriel, Bharambe, Eisenman, Yazdan, James, Maurer, Leonhardi, Huang, Loyd, Paola, Paranjape, Liu, Wu, Ni, Hancock, Wasti, Spence, Stojkovic, Gamido, Montalvo, Parker, Burton, Mejia, Liu, Wang, Kim, Zhou, Hu, Chu, Cai, Tindal, Feichtenhofer, Gao, Civin, Beaty, Kreymer, Li, Adkins, Xu, Testuggine, David, Parikh, Liskovich, Foss, Wang, Le, Holland, Dowling, Jamil, Montgomery, Presani, Hahn, Wood, Le, Brinkman, Arcaute, Dunbar, Smothers, Sun, Kreuk, Tian, Kokkinos, Ozgenel, Caggioni, Kanayet, Seide, Florez, Schwarz, Badeer, Swee, Halpern, Herman, Sizov, Guangyi, Zhang, Lakshminarayanan, Inan, Shojanazeri, Zou, Wang, Zha, Habeeb, Rudolph, Suk, Aspegren, Goldman, Zhan, Damlaj, Molybog, Tufanov, Leontiadis, Veliche, Gat, Weissman, Geboski, Kohli, Lam, Asher, Gaya, Marcus, Tang, Chan, Zhen, Reizenstein, Teboul, Zhong, Jin, Yang, Cummings, Carvill, Shepard, McPhie,
  Torres, Ginsburg, Wang, Wu, U, Saxena, Khandelwal, Zand, Matosich, Veeraraghavan, Michelena, Li, Jagadeesh, Huang, Chawla, Huang, Chen, Garg, A, Silva, Bell, Zhang, Guo, Yu, Moshkovich, Wehrstedt, Khabsa, Avalani, Bhatt, Mankus, Hasson, Lennie, Reso, Groshev, Naumov, Lathi, Keneally, Liu, Seltzer, Valko, Restrepo, Patel, Vyatskov, Samvelyan, Clark, Macey, Wang, Hermoso, Metanat, Rastegari, Bansal, Santhanam, Parks, White, Bawa, Singhal, Egebo, Usunier, Mehta, Laptev, Dong, Cheng, Chernoguz, Hart, Salpekar, Kalinli, Kent, Parekh, Saab, Balaji, Rittner, Bontrager, Roux, Dollar, Zvyagina, Ratanchandani, Yuvraj, Liang, Alao, Rodriguez, Ayub, Murthy, Nayani, Mitra, Parthasarathy, Li, Hogan, Battey, Wang, Howes, Rinott, Mehta, Siby, Bondu, Datta, Chugh, Hunt, Dhillon, Sidorov, Pan, Mahajan, Verma, Yamamoto, Ramaswamy, Lindsay, Lindsay, Feng, Lin, Zha, Patil, Shankar, Zhang, Zhang, Wang, Agarwal, Sajuyigbe, Chintala, Max, Chen, Kehoe, Satterfield, Govindaprasad, Gupta, Deng, Cho, Virk, Subramanian, Choudhury,
  Goldman, Remez, Glaser, Best, Koehler, Robinson, Li, Zhang, Matthews, Chou, Shaked, Vontimitta, Ajayi, Montanez, Mohan, Kumar, Mangla, Ionescu, Poenaru, Mihailescu, Ivanov, Li, Wang, Jiang, Bouaziz, Constable, Tang, Wu, Wang, Wu, Gao, Kleinman, Chen, Hu, Jia, Qi, Li, Zhang, Zhang, Adi, Nam, Yu, Wang, Zhao, Hao, Qian, Li, He, Rait, DeVito, Rosnbrick, Wen, Yang, Zhao, and Ma}]{grattafiori2024llama3herdmodels}
Aaron Grattafiori, Abhimanyu Dubey, Abhinav Jauhri, Abhinav Pandey, Abhishek Kadian, Ahmad Al-Dahle, Aiesha Letman, Akhil Mathur, Alan Schelten, Alex Vaughan, Amy Yang, Angela Fan, Anirudh Goyal, Anthony Hartshorn, Aobo Yang, Archi Mitra, Archie Sravankumar, Artem Korenev, Arthur Hinsvark, Arun Rao, Aston Zhang, Aurelien Rodriguez, Austen Gregerson, Ava Spataru, Baptiste Roziere, Bethany Biron, Binh Tang, Bobbie Chern, Charlotte Caucheteux, Chaya Nayak, Chloe Bi, Chris Marra, Chris McConnell, Christian Keller, Christophe Touret, Chunyang Wu, Corinne Wong, Cristian~Canton Ferrer, Cyrus Nikolaidis, Damien Allonsius, Daniel Song, Danielle Pintz, Danny Livshits, Danny Wyatt, David Esiobu, Dhruv Choudhary, Dhruv Mahajan, Diego Garcia-Olano, Diego Perino, Dieuwke Hupkes, Egor Lakomkin, Ehab AlBadawy, Elina Lobanova, Emily Dinan, Eric~Michael Smith, Filip Radenovic, Francisco Guzmán, Frank Zhang, Gabriel Synnaeve, Gabrielle Lee, Georgia~Lewis Anderson, Govind Thattai, Graeme Nail, Gregoire Mialon, Guan Pang,
  Guillem Cucurell, Hailey Nguyen, Hannah Korevaar, Hu~Xu, Hugo Touvron, Iliyan Zarov, Imanol~Arrieta Ibarra, Isabel Kloumann, Ishan Misra, Ivan Evtimov, Jack Zhang, Jade Copet, Jaewon Lee, Jan Geffert, Jana Vranes, Jason Park, Jay Mahadeokar, Jeet Shah, Jelmer van~der Linde, Jennifer Billock, Jenny Hong, Jenya Lee, Jeremy Fu, Jianfeng Chi, Jianyu Huang, Jiawen Liu, Jie Wang, Jiecao Yu, Joanna Bitton, Joe Spisak, Jongsoo Park, Joseph Rocca, Joshua Johnstun, Joshua Saxe, Junteng Jia, Kalyan~Vasuden Alwala, Karthik Prasad, Kartikeya Upasani, Kate Plawiak, Ke~Li, Kenneth Heafield, Kevin Stone, Khalid El-Arini, Krithika Iyer, Kshitiz Malik, Kuenley Chiu, Kunal Bhalla, Kushal Lakhotia, Lauren Rantala-Yeary, Laurens van~der Maaten, Lawrence Chen, Liang Tan, Liz Jenkins, Louis Martin, Lovish Madaan, Lubo Malo, Lukas Blecher, Lukas Landzaat, Luke de~Oliveira, Madeline Muzzi, Mahesh Pasupuleti, Mannat Singh, Manohar Paluri, Marcin Kardas, Maria Tsimpoukelli, Mathew Oldham, Mathieu Rita, Maya Pavlova, Melanie Kambadur,
  Mike Lewis, Min Si, Mitesh~Kumar Singh, Mona Hassan, Naman Goyal, Narjes Torabi, Nikolay Bashlykov, Nikolay Bogoychev, Niladri Chatterji, Ning Zhang, Olivier Duchenne, Onur Çelebi, Patrick Alrassy, Pengchuan Zhang, Pengwei Li, Petar Vasic, Peter Weng, Prajjwal Bhargava, Pratik Dubal, Praveen Krishnan, Punit~Singh Koura, Puxin Xu, Qing He, Qingxiao Dong, Ragavan Srinivasan, Raj Ganapathy, Ramon Calderer, Ricardo~Silveira Cabral, Robert Stojnic, Roberta Raileanu, Rohan Maheswari, Rohit Girdhar, Rohit Patel, Romain Sauvestre, Ronnie Polidoro, Roshan Sumbaly, Ross Taylor, Ruan Silva, Rui Hou, Rui Wang, Saghar Hosseini, Sahana Chennabasappa, Sanjay Singh, Sean Bell, Seohyun~Sonia Kim, Sergey Edunov, Shaoliang Nie, Sharan Narang, Sharath Raparthy, Sheng Shen, Shengye Wan, Shruti Bhosale, Shun Zhang, Simon Vandenhende, Soumya Batra, Spencer Whitman, Sten Sootla, Stephane Collot, Suchin Gururangan, Sydney Borodinsky, Tamar Herman, Tara Fowler, Tarek Sheasha, Thomas Georgiou, Thomas Scialom, Tobias Speckbacher,
  Todor Mihaylov, Tong Xiao, Ujjwal Karn, Vedanuj Goswami, Vibhor Gupta, Vignesh Ramanathan, Viktor Kerkez, Vincent Gonguet, Virginie Do, Vish Vogeti, Vítor Albiero, Vladan Petrovic, Weiwei Chu, Wenhan Xiong, Wenyin Fu, Whitney Meers, Xavier Martinet, Xiaodong Wang, Xiaofang Wang, Xiaoqing~Ellen Tan, Xide Xia, Xinfeng Xie, Xuchao Jia, Xuewei Wang, Yaelle Goldschlag, Yashesh Gaur, Yasmine Babaei, Yi~Wen, Yiwen Song, Yuchen Zhang, Yue Li, Yuning Mao, Zacharie~Delpierre Coudert, Zheng Yan, Zhengxing Chen, Zoe Papakipos, Aaditya Singh, Aayushi Srivastava, Abha Jain, Adam Kelsey, Adam Shajnfeld, Adithya Gangidi, Adolfo Victoria, Ahuva Goldstand, Ajay Menon, Ajay Sharma, Alex Boesenberg, Alexei Baevski, Allie Feinstein, Amanda Kallet, Amit Sangani, Amos Teo, Anam Yunus, Andrei Lupu, Andres Alvarado, Andrew Caples, Andrew Gu, Andrew Ho, Andrew Poulton, Andrew Ryan, Ankit Ramchandani, Annie Dong, Annie Franco, Anuj Goyal, Aparajita Saraf, Arkabandhu Chowdhury, Ashley Gabriel, Ashwin Bharambe, Assaf Eisenman, Azadeh
  Yazdan, Beau James, Ben Maurer, Benjamin Leonhardi, Bernie Huang, Beth Loyd, Beto~De Paola, Bhargavi Paranjape, Bing Liu, Bo~Wu, Boyu Ni, Braden Hancock, Bram Wasti, Brandon Spence, Brani Stojkovic, Brian Gamido, Britt Montalvo, Carl Parker, Carly Burton, Catalina Mejia, Ce~Liu, Changhan Wang, Changkyu Kim, Chao Zhou, Chester Hu, Ching-Hsiang Chu, Chris Cai, Chris Tindal, Christoph Feichtenhofer, Cynthia Gao, Damon Civin, Dana Beaty, Daniel Kreymer, Daniel Li, David Adkins, David Xu, Davide Testuggine, Delia David, Devi Parikh, Diana Liskovich, Didem Foss, Dingkang Wang, Duc Le, Dustin Holland, Edward Dowling, Eissa Jamil, Elaine Montgomery, Eleonora Presani, Emily Hahn, Emily Wood, Eric-Tuan Le, Erik Brinkman, Esteban Arcaute, Evan Dunbar, Evan Smothers, Fei Sun, Felix Kreuk, Feng Tian, Filippos Kokkinos, Firat Ozgenel, Francesco Caggioni, Frank Kanayet, Frank Seide, Gabriela~Medina Florez, Gabriella Schwarz, Gada Badeer, Georgia Swee, Gil Halpern, Grant Herman, Grigory Sizov, Guangyi, Zhang, Guna
  Lakshminarayanan, Hakan Inan, Hamid Shojanazeri, Han Zou, Hannah Wang, Hanwen Zha, Haroun Habeeb, Harrison Rudolph, Helen Suk, Henry Aspegren, Hunter Goldman, Hongyuan Zhan, Ibrahim Damlaj, Igor Molybog, Igor Tufanov, Ilias Leontiadis, Irina-Elena Veliche, Itai Gat, Jake Weissman, James Geboski, James Kohli, Janice Lam, Japhet Asher, Jean-Baptiste Gaya, Jeff Marcus, Jeff Tang, Jennifer Chan, Jenny Zhen, Jeremy Reizenstein, Jeremy Teboul, Jessica Zhong, Jian Jin, Jingyi Yang, Joe Cummings, Jon Carvill, Jon Shepard, Jonathan McPhie, Jonathan Torres, Josh Ginsburg, Junjie Wang, Kai Wu, Kam~Hou U, Karan Saxena, Kartikay Khandelwal, Katayoun Zand, Kathy Matosich, Kaushik Veeraraghavan, Kelly Michelena, Keqian Li, Kiran Jagadeesh, Kun Huang, Kunal Chawla, Kyle Huang, Lailin Chen, Lakshya Garg, Lavender A, Leandro Silva, Lee Bell, Lei Zhang, Liangpeng Guo, Licheng Yu, Liron Moshkovich, Luca Wehrstedt, Madian Khabsa, Manav Avalani, Manish Bhatt, Martynas Mankus, Matan Hasson, Matthew Lennie, Matthias Reso, Maxim
  Groshev, Maxim Naumov, Maya Lathi, Meghan Keneally, Miao Liu, Michael~L. Seltzer, Michal Valko, Michelle Restrepo, Mihir Patel, Mik Vyatskov, Mikayel Samvelyan, Mike Clark, Mike Macey, Mike Wang, Miquel~Jubert Hermoso, Mo~Metanat, Mohammad Rastegari, Munish Bansal, Nandhini Santhanam, Natascha Parks, Natasha White, Navyata Bawa, Nayan Singhal, Nick Egebo, Nicolas Usunier, Nikhil Mehta, Nikolay~Pavlovich Laptev, Ning Dong, Norman Cheng, Oleg Chernoguz, Olivia Hart, Omkar Salpekar, Ozlem Kalinli, Parkin Kent, Parth Parekh, Paul Saab, Pavan Balaji, Pedro Rittner, Philip Bontrager, Pierre Roux, Piotr Dollar, Polina Zvyagina, Prashant Ratanchandani, Pritish Yuvraj, Qian Liang, Rachad Alao, Rachel Rodriguez, Rafi Ayub, Raghotham Murthy, Raghu Nayani, Rahul Mitra, Rangaprabhu Parthasarathy, Raymond Li, Rebekkah Hogan, Robin Battey, Rocky Wang, Russ Howes, Ruty Rinott, Sachin Mehta, Sachin Siby, Sai~Jayesh Bondu, Samyak Datta, Sara Chugh, Sara Hunt, Sargun Dhillon, Sasha Sidorov, Satadru Pan, Saurabh Mahajan,
  Saurabh Verma, Seiji Yamamoto, Sharadh Ramaswamy, Shaun Lindsay, Shaun Lindsay, Sheng Feng, Shenghao Lin, Shengxin~Cindy Zha, Shishir Patil, Shiva Shankar, Shuqiang Zhang, Shuqiang Zhang, Sinong Wang, Sneha Agarwal, Soji Sajuyigbe, Soumith Chintala, Stephanie Max, Stephen Chen, Steve Kehoe, Steve Satterfield, Sudarshan Govindaprasad, Sumit Gupta, Summer Deng, Sungmin Cho, Sunny Virk, Suraj Subramanian, Sy~Choudhury, Sydney Goldman, Tal Remez, Tamar Glaser, Tamara Best, Thilo Koehler, Thomas Robinson, Tianhe Li, Tianjun Zhang, Tim Matthews, Timothy Chou, Tzook Shaked, Varun Vontimitta, Victoria Ajayi, Victoria Montanez, Vijai Mohan, Vinay~Satish Kumar, Vishal Mangla, Vlad Ionescu, Vlad Poenaru, Vlad~Tiberiu Mihailescu, Vladimir Ivanov, Wei Li, Wenchen Wang, Wenwen Jiang, Wes Bouaziz, Will Constable, Xiaocheng Tang, Xiaojian Wu, Xiaolan Wang, Xilun Wu, Xinbo Gao, Yaniv Kleinman, Yanjun Chen, Ye~Hu, Ye~Jia, Ye~Qi, Yenda Li, Yilin Zhang, Ying Zhang, Yossi Adi, Youngjin Nam, Yu, Wang, Yu~Zhao, Yuchen Hao, Yundi
  Qian, Yunlu Li, Yuzi He, Zach Rait, Zachary DeVito, Zef Rosnbrick, Zhaoduo Wen, Zhenyu Yang, Zhiwei Zhao, and Zhiyu Ma. 2024.
\newblock \href {https://arxiv.org/abs/2407.21783} {The llama 3 herd of models}.
\newblock \emph{Preprint}, arXiv:2407.21783.

\bibitem[{Hopcroft et~al.(2001)Hopcroft, Motwani, and Ullman}]{hopcroft2001introduction}
John~E Hopcroft, Rajeev Motwani, and Jeffrey~D Ullman. 2001.
\newblock Introduction to automata theory, languages, and computation.
\newblock \emph{Acm Sigact News}, 32(1):60--65.

\bibitem[{Horning(1969)}]{horning1969study}
James~Jay Horning. 1969.
\newblock \emph{A study of grammatical inference}.
\newblock Stanford University.

\bibitem[{Huang et~al.(2023)Huang, Bu, Zhang, Luck, and Cui}]{Huang2023AgentCoder}
Dong Huang, Qingwen Bu, Jie~M. Zhang, Michael Luck, and Heming Cui. 2023.
\newblock \href {https://doi.org/10.48550/arXiv.2312.13010} {Agentcoder: Multi-agent-based code generation with iterative testing and optimisation}.
\newblock \emph{ArXiv}, abs/2312.13010.

\bibitem[{Imani et~al.(2023)Imani, Du, and Shrivastava}]{imani-etal-2023-mathprompter}
Shima Imani, Liang Du, and Harsh Shrivastava. 2023.
\newblock \href {https://doi.org/10.18653/v1/2023.acl-industry.4} {{M}ath{P}rompter: Mathematical reasoning using large language models}.
\newblock In \emph{Proceedings of the 61st Annual Meeting of the Association for Computational Linguistics (Volume 5: Industry Track)}, pages 37--42, Toronto, Canada. Association for Computational Linguistics.

\bibitem[{Jiang et~al.(2023{\natexlab{a}})Jiang, Sablayrolles, Mensch, Bamford, Chaplot, de~las Casas, Bressand, Lengyel, Lample, Saulnier, Lavaud, Lachaux, Stock, Scao, Lavril, Wang, Lacroix, and Sayed}]{jiang2023mistral7b}
Albert~Q. Jiang, Alexandre Sablayrolles, Arthur Mensch, Chris Bamford, Devendra~Singh Chaplot, Diego de~las Casas, Florian Bressand, Gianna Lengyel, Guillaume Lample, Lucile Saulnier, Lélio~Renard Lavaud, Marie-Anne Lachaux, Pierre Stock, Teven~Le Scao, Thibaut Lavril, Thomas Wang, Timothée Lacroix, and William~El Sayed. 2023{\natexlab{a}}.
\newblock \href {https://arxiv.org/abs/2310.06825} {Mistral 7b}.
\newblock \emph{Preprint}, arXiv:2310.06825.

\bibitem[{Jiang et~al.(2024{\natexlab{a}})Jiang, Wang, Shen, Kim, and Kim}]{jiang2024surveylargelanguagemodels}
Juyong Jiang, Fan Wang, Jiasi Shen, Sungju Kim, and Sunghun Kim. 2024{\natexlab{a}}.
\newblock \href {https://arxiv.org/abs/2406.00515} {A survey on large language models for code generation}.
\newblock \emph{Preprint}, arXiv:2406.00515.

\bibitem[{Jiang et~al.(2024{\natexlab{b}})Jiang, Wang, Shen, Kim, and Kim}]{Jiang2024A}
Juyong Jiang, Fan Wang, Jiasi Shen, Sungju Kim, and Sunghun Kim. 2024{\natexlab{b}}.
\newblock \href {https://doi.org/10.48550/arXiv.2406.00515} {A survey on large language models for code generation}.
\newblock \emph{ArXiv}, abs/2406.00515.

\bibitem[{Jiang et~al.(2023{\natexlab{b}})Jiang, Dong, Wang, Shang, and Li}]{Jiang2023SelfPlanning}
Xue Jiang, Yihong Dong, Lecheng Wang, Qiwei Shang, and Ge~Li. 2023{\natexlab{b}}.
\newblock \href {https://doi.org/10.1145/3672456} {Self-planning code generation with large language models}.
\newblock \emph{ACM Transactions on Software Engineering and Methodology}.

\bibitem[{Johnson and Hill(1978)}]{Johnson1978YaccYA}
S.~C. Johnson and Murray Hill. 1978.
\newblock \href {https://api.semanticscholar.org/CorpusID:62540186} {Yacc: Yet another compiler-compiler}.

\bibitem[{Kai et~al.(2024)Kai, Hou, Huang, and Lin}]{kai2024leveraginggrammarinductionlanguage}
Jushi Kai, Shengyuan Hou, Yusheng Huang, and Zhouhan Lin. 2024.
\newblock \href {https://arxiv.org/abs/2410.04878} {Leveraging grammar induction for language understanding and generation}.
\newblock \emph{Preprint}, arXiv:2410.04878.

\bibitem[{Li et~al.(2023)Li, Mora, Polgreen, and Seshia}]{ga-metagrammar}
Yixuan Li, Federico Mora, Elizabeth Polgreen, and Sanjit~A Seshia. 2023.
\newblock Genetic algorithms for searching a matrix of metagrammars for synthesis.
\newblock \emph{arXiv preprint arXiv:2306.00521}.

\bibitem[{Li et~al.(2024)Li, Parsert, and Polgreen}]{LLM-SYGUS}
Yixuan Li, Julian Parsert, and Elizabeth Polgreen. 2024.
\newblock \href {https://doi.org/10.1007/978-3-031-65630-9_15} {Guiding enumerative program synthesis with large language models}.
\newblock In \emph{Computer Aided Verification}, CAV 2024, pages 280--301, Cham. Springer Nature Switzerland.

\bibitem[{Lozhkov et~al.(2024)Lozhkov, Li, Allal, Cassano, Lamy-Poirier, Tazi, Tang, Pykhtar, Liu, Wei, Liu, Tian, Kocetkov, Zucker, Belkada, Wang, Liu, Abulkhanov, Paul, Li, Li, Risdal, Li, Zhu, Zhuo, Zheltonozhskii, Dade, Yu, Krauß, Jain, Su, He, Dey, Abati, Chai, Muennighoff, Tang, Oblokulov, Akiki, Marone, Mou, Mishra, Gu, Hui, Dao, Zebaze, Dehaene, Patry, Xu, McAuley, Hu, Scholak, Paquet, Robinson, Anderson, Chapados, Patwary, Tajbakhsh, Jernite, Ferrandis, Zhang, Hughes, Wolf, Guha, von Werra, and de~Vries}]{lozhkov2024starcoder2stackv2}
Anton Lozhkov, Raymond Li, Loubna~Ben Allal, Federico Cassano, Joel Lamy-Poirier, Nouamane Tazi, Ao~Tang, Dmytro Pykhtar, Jiawei Liu, Yuxiang Wei, Tianyang Liu, Max Tian, Denis Kocetkov, Arthur Zucker, Younes Belkada, Zijian Wang, Qian Liu, Dmitry Abulkhanov, Indraneil Paul, Zhuang Li, Wen-Ding Li, Megan Risdal, Jia Li, Jian Zhu, Terry~Yue Zhuo, Evgenii Zheltonozhskii, Nii Osae~Osae Dade, Wenhao Yu, Lucas Krauß, Naman Jain, Yixuan Su, Xuanli He, Manan Dey, Edoardo Abati, Yekun Chai, Niklas Muennighoff, Xiangru Tang, Muhtasham Oblokulov, Christopher Akiki, Marc Marone, Chenghao Mou, Mayank Mishra, Alex Gu, Binyuan Hui, Tri Dao, Armel Zebaze, Olivier Dehaene, Nicolas Patry, Canwen Xu, Julian McAuley, Han Hu, Torsten Scholak, Sebastien Paquet, Jennifer Robinson, Carolyn~Jane Anderson, Nicolas Chapados, Mostofa Patwary, Nima Tajbakhsh, Yacine Jernite, Carlos~Muñoz Ferrandis, Lingming Zhang, Sean Hughes, Thomas Wolf, Arjun Guha, Leandro von Werra, and Harm de~Vries. 2024.
\newblock \href {https://arxiv.org/abs/2402.19173} {Starcoder 2 and the stack v2: The next generation}.
\newblock \emph{Preprint}, arXiv:2402.19173.

\bibitem[{Madaan et~al.(2023)Madaan, Tandon, Gupta, Hallinan, Gao, Wiegreffe, Alon, Dziri, Prabhumoye, Yang, Gupta, Majumder, Hermann, Welleck, Yazdanbakhsh, and Clark}]{madaan2023selfrefine}
Aman Madaan, Niket Tandon, Prakhar Gupta, Skyler Hallinan, Luyu Gao, Sarah Wiegreffe, Uri Alon, Nouha Dziri, Shrimai Prabhumoye, Yiming Yang, Shashank Gupta, Bodhisattwa~Prasad Majumder, Katherine Hermann, Sean Welleck, Amir Yazdanbakhsh, and Peter Clark. 2023.
\newblock \href {https://openreview.net/forum?id=S37hOerQLB} {Self-refine: Iterative refinement with self-feedback}.
\newblock In \emph{Thirty-seventh Conference on Neural Information Processing Systems}.

\bibitem[{MistralAI(2024)}]{mistral2024codestral}
MistralAI. 2024.
\newblock Codestral.
\newblock \url{https://mistral.ai/news/codestral/}.

\bibitem[{OpenAI et~al.(2024)OpenAI, Achiam, Adler, Agarwal, Ahmad, Akkaya, Aleman, Almeida, Altenschmidt, Altman, Anadkat, Avila, Babuschkin, Balaji, Balcom, Baltescu, Bao, Bavarian, Belgum, Bello, Berdine, Bernadett-Shapiro, Berner, Bogdonoff, Boiko, Boyd, Brakman, Brockman, Brooks, Brundage, Button, Cai, Campbell, Cann, Carey, Carlson, Carmichael, Chan, Chang, Chantzis, Chen, Chen, Chen, Chen, Chen, Chess, Cho, Chu, Chung, Cummings, Currier, Dai, Decareaux, Degry, Deutsch, Deville, Dhar, Dohan, Dowling, Dunning, Ecoffet, Eleti, Eloundou, Farhi, Fedus, Felix, Fishman, Forte, Fulford, Gao, Georges, Gibson, Goel, Gogineni, Goh, Gontijo-Lopes, Gordon, Grafstein, Gray, Greene, Gross, Gu, Guo, Hallacy, Han, Harris, He, Heaton, Heidecke, Hesse, Hickey, Hickey, Hoeschele, Houghton, Hsu, Hu, Hu, Huizinga, Jain, Jain, Jang, Jiang, Jiang, Jin, Jin, Jomoto, Jonn, Jun, Kaftan, Łukasz Kaiser, Kamali, Kanitscheider, Keskar, Khan, Kilpatrick, Kim, Kim, Kim, Kirchner, Kiros, Knight, Kokotajlo, Łukasz Kondraciuk,
  Kondrich, Konstantinidis, Kosic, Krueger, Kuo, Lampe, Lan, Lee, Leike, Leung, Levy, Li, Lim, Lin, Lin, Litwin, Lopez, Lowe, Lue, Makanju, Malfacini, Manning, Markov, Markovski, Martin, Mayer, Mayne, McGrew, McKinney, McLeavey, McMillan, McNeil, Medina, Mehta, Menick, Metz, Mishchenko, Mishkin, Monaco, Morikawa, Mossing, Mu, Murati, Murk, Mély, Nair, Nakano, Nayak, Neelakantan, Ngo, Noh, Ouyang, O'Keefe, Pachocki, Paino, Palermo, Pantuliano, Parascandolo, Parish, Parparita, Passos, Pavlov, Peng, Perelman, de~Avila Belbute~Peres, Petrov, de~Oliveira~Pinto, Michael, Pokorny, Pokrass, Pong, Powell, Power, Power, Proehl, Puri, Radford, Rae, Ramesh, Raymond, Real, Rimbach, Ross, Rotsted, Roussez, Ryder, Saltarelli, Sanders, Santurkar, Sastry, Schmidt, Schnurr, Schulman, Selsam, Sheppard, Sherbakov, Shieh, Shoker, Shyam, Sidor, Sigler, Simens, Sitkin, Slama, Sohl, Sokolowsky, Song, Staudacher, Such, Summers, Sutskever, Tang, Tezak, Thompson, Tillet, Tootoonchian, Tseng, Tuggle, Turley, Tworek, Uribe, Vallone,
  Vijayvergiya, Voss, Wainwright, Wang, Wang, Wang, Ward, Wei, Weinmann, Welihinda, Welinder, Weng, Weng, Wiethoff, Willner, Winter, Wolrich, Wong, Workman, Wu, Wu, Wu, Xiao, Xu, Yoo, Yu, Yuan, Zaremba, Zellers, Zhang, Zhang, Zhao, Zheng, Zhuang, Zhuk, and Zoph}]{openai2024gpt4technicalreport}
OpenAI, Josh Achiam, Steven Adler, Sandhini Agarwal, Lama Ahmad, Ilge Akkaya, Florencia~Leoni Aleman, Diogo Almeida, Janko Altenschmidt, Sam Altman, Shyamal Anadkat, Red Avila, Igor Babuschkin, Suchir Balaji, Valerie Balcom, Paul Baltescu, Haiming Bao, Mohammad Bavarian, Jeff Belgum, Irwan Bello, Jake Berdine, Gabriel Bernadett-Shapiro, Christopher Berner, Lenny Bogdonoff, Oleg Boiko, Madelaine Boyd, Anna-Luisa Brakman, Greg Brockman, Tim Brooks, Miles Brundage, Kevin Button, Trevor Cai, Rosie Campbell, Andrew Cann, Brittany Carey, Chelsea Carlson, Rory Carmichael, Brooke Chan, Che Chang, Fotis Chantzis, Derek Chen, Sully Chen, Ruby Chen, Jason Chen, Mark Chen, Ben Chess, Chester Cho, Casey Chu, Hyung~Won Chung, Dave Cummings, Jeremiah Currier, Yunxing Dai, Cory Decareaux, Thomas Degry, Noah Deutsch, Damien Deville, Arka Dhar, David Dohan, Steve Dowling, Sheila Dunning, Adrien Ecoffet, Atty Eleti, Tyna Eloundou, David Farhi, Liam Fedus, Niko Felix, Simón~Posada Fishman, Juston Forte, Isabella Fulford, Leo
  Gao, Elie Georges, Christian Gibson, Vik Goel, Tarun Gogineni, Gabriel Goh, Rapha Gontijo-Lopes, Jonathan Gordon, Morgan Grafstein, Scott Gray, Ryan Greene, Joshua Gross, Shixiang~Shane Gu, Yufei Guo, Chris Hallacy, Jesse Han, Jeff Harris, Yuchen He, Mike Heaton, Johannes Heidecke, Chris Hesse, Alan Hickey, Wade Hickey, Peter Hoeschele, Brandon Houghton, Kenny Hsu, Shengli Hu, Xin Hu, Joost Huizinga, Shantanu Jain, Shawn Jain, Joanne Jang, Angela Jiang, Roger Jiang, Haozhun Jin, Denny Jin, Shino Jomoto, Billie Jonn, Heewoo Jun, Tomer Kaftan, Łukasz Kaiser, Ali Kamali, Ingmar Kanitscheider, Nitish~Shirish Keskar, Tabarak Khan, Logan Kilpatrick, Jong~Wook Kim, Christina Kim, Yongjik Kim, Jan~Hendrik Kirchner, Jamie Kiros, Matt Knight, Daniel Kokotajlo, Łukasz Kondraciuk, Andrew Kondrich, Aris Konstantinidis, Kyle Kosic, Gretchen Krueger, Vishal Kuo, Michael Lampe, Ikai Lan, Teddy Lee, Jan Leike, Jade Leung, Daniel Levy, Chak~Ming Li, Rachel Lim, Molly Lin, Stephanie Lin, Mateusz Litwin, Theresa Lopez, Ryan
  Lowe, Patricia Lue, Anna Makanju, Kim Malfacini, Sam Manning, Todor Markov, Yaniv Markovski, Bianca Martin, Katie Mayer, Andrew Mayne, Bob McGrew, Scott~Mayer McKinney, Christine McLeavey, Paul McMillan, Jake McNeil, David Medina, Aalok Mehta, Jacob Menick, Luke Metz, Andrey Mishchenko, Pamela Mishkin, Vinnie Monaco, Evan Morikawa, Daniel Mossing, Tong Mu, Mira Murati, Oleg Murk, David Mély, Ashvin Nair, Reiichiro Nakano, Rajeev Nayak, Arvind Neelakantan, Richard Ngo, Hyeonwoo Noh, Long Ouyang, Cullen O'Keefe, Jakub Pachocki, Alex Paino, Joe Palermo, Ashley Pantuliano, Giambattista Parascandolo, Joel Parish, Emy Parparita, Alex Passos, Mikhail Pavlov, Andrew Peng, Adam Perelman, Filipe de~Avila Belbute~Peres, Michael Petrov, Henrique~Ponde de~Oliveira~Pinto, Michael, Pokorny, Michelle Pokrass, Vitchyr~H. Pong, Tolly Powell, Alethea Power, Boris Power, Elizabeth Proehl, Raul Puri, Alec Radford, Jack Rae, Aditya Ramesh, Cameron Raymond, Francis Real, Kendra Rimbach, Carl Ross, Bob Rotsted, Henri Roussez,
  Nick Ryder, Mario Saltarelli, Ted Sanders, Shibani Santurkar, Girish Sastry, Heather Schmidt, David Schnurr, John Schulman, Daniel Selsam, Kyla Sheppard, Toki Sherbakov, Jessica Shieh, Sarah Shoker, Pranav Shyam, Szymon Sidor, Eric Sigler, Maddie Simens, Jordan Sitkin, Katarina Slama, Ian Sohl, Benjamin Sokolowsky, Yang Song, Natalie Staudacher, Felipe~Petroski Such, Natalie Summers, Ilya Sutskever, Jie Tang, Nikolas Tezak, Madeleine~B. Thompson, Phil Tillet, Amin Tootoonchian, Elizabeth Tseng, Preston Tuggle, Nick Turley, Jerry Tworek, Juan Felipe~Cerón Uribe, Andrea Vallone, Arun Vijayvergiya, Chelsea Voss, Carroll Wainwright, Justin~Jay Wang, Alvin Wang, Ben Wang, Jonathan Ward, Jason Wei, CJ~Weinmann, Akila Welihinda, Peter Welinder, Jiayi Weng, Lilian Weng, Matt Wiethoff, Dave Willner, Clemens Winter, Samuel Wolrich, Hannah Wong, Lauren Workman, Sherwin Wu, Jeff Wu, Michael Wu, Kai Xiao, Tao Xu, Sarah Yoo, Kevin Yu, Qiming Yuan, Wojciech Zaremba, Rowan Zellers, Chong Zhang, Marvin Zhang, Shengjia
  Zhao, Tianhao Zheng, Juntang Zhuang, William Zhuk, and Barret Zoph. 2024.
\newblock \href {https://arxiv.org/abs/2303.08774} {Gpt-4 technical report}.
\newblock \emph{Preprint}, arXiv:2303.08774.

\bibitem[{Pan et~al.(2023)Pan, Albalak, Wang, and Wang}]{Pan_Albalak_Wang_Wang_2023}
Liangming Pan, Alon Albalak, Xinyi Wang, and William~Yang Wang. 2023.
\newblock \href {http://arxiv.org/abs/2305.12295} {Logic-lm: Empowering large language models with symbolic solvers for faithful logical reasoning}.
\newblock (arXiv:2305.12295).
\newblock ArXiv:2305.12295 [cs].

\bibitem[{Parr(2013)}]{ANTLR4}
Terence Parr. 2013.
\newblock \emph{The Definitive ANTLR 4 Reference}, 2nd edition.
\newblock Pragmatic Bookshelf.

\bibitem[{Pedro et~al.(2013)Pedro, Nunes, and Machado-Lima}]{Pedro2013Using}
Ricardo Wandré~Dias Pedro, Fátima L.~S. Nunes, and Ariane Machado-Lima. 2013.
\newblock \href {https://doi.org/10.1145/2543581.2543593} {Using grammars for pattern recognition in images}.
\newblock \emph{ACM Computing Surveys (CSUR)}, 46:1 -- 34.

\bibitem[{Richetin and Vernadat(1984)}]{Richetin1984Efficient}
M.~Richetin and F.~Vernadat. 1984.
\newblock \href {https://doi.org/10.1016/0031-3203(84)90063-3} {Efficient regular grammatical inference for pattern recognition}.
\newblock \emph{Pattern Recognit.}, 17:245--250.

\bibitem[{Rodrigues and Lopes(2007)}]{Rodrigues2007GeneticPF}
Ernesto Rodrigues and Heitor~Silv{\'e}rio Lopes. 2007.
\newblock \href {https://api.semanticscholar.org/CorpusID:255763} {Genetic programming for induction of context-free grammars}.
\newblock \emph{Seventh International Conference on Intelligent Systems Design and Applications (ISDA 2007)}, pages 297--302.

\bibitem[{Schröder and Cito(2022)}]{Schröder_Cito_2022}
Michael Schröder and Jürgen Cito. 2022.
\newblock \href {https://doi.org/10.1145/3510455.3512787} {Grammars for free: toward grammar inference for ad hoc parsers}.
\newblock In \emph{Proceedings of the ACM/IEEE 44th International Conference on Software Engineering: New Ideas and Emerging Results}, page 41–45, Pittsburgh Pennsylvania. ACM.

\bibitem[{Shinn et~al.(2023)Shinn, Cassano, Gopinath, Narasimhan, and Yao}]{shinn2023reflexion}
Noah Shinn, Federico Cassano, Ashwin Gopinath, Karthik~R Narasimhan, and Shunyu Yao. 2023.
\newblock \href {https://openreview.net/forum?id=vAElhFcKW6} {Reflexion: language agents with verbal reinforcement learning}.
\newblock In \emph{Thirty-seventh Conference on Neural Information Processing Systems}.

\bibitem[{Stevenson and Cordy(2014{\natexlab{a}})}]{Stevenson2014A}
Andrew Stevenson and J.~Cordy. 2014{\natexlab{a}}.
\newblock \href {https://doi.org/10.1016/j.scico.2014.05.008} {A survey of grammatical inference in software engineering}.
\newblock \emph{Sci. Comput. Program.}, 96:444--459.

\bibitem[{Stevenson and Cordy(2014{\natexlab{b}})}]{Stevenson_Cordy_2014}
Andrew Stevenson and James~R. Cordy. 2014{\natexlab{b}}.
\newblock \href {https://doi.org/10.1016/j.scico.2014.05.008} {A survey of grammatical inference in software engineering}.
\newblock \emph{Science of Computer Programming}, 96:444–459.

\bibitem[{Tang and Belle(2024)}]{tang2024tom}
Weizhi Tang and Vaishak Belle. 2024.
\newblock Tom-lm: Delegating theory of mind reasoning to external symbolic executors in large language models.
\newblock \emph{arXiv preprint arXiv:2404.15515}.

\bibitem[{Team et~al.(2024)Team, Riviere, Pathak, Sessa, Hardin, Bhupatiraju, Hussenot, Mesnard, Shahriari, Ramé, Ferret, Liu, Tafti, Friesen, Casbon, Ramos, Kumar, Lan, Jerome, Tsitsulin, Vieillard, Stanczyk, Girgin, Momchev, Hoffman, Thakoor, Grill, Neyshabur, Bachem, Walton, Severyn, Parrish, Ahmad, Hutchison, Abdagic, Carl, Shen, Brock, Coenen, Laforge, Paterson, Bastian, Piot, Wu, Royal, Chen, Kumar, Perry, Welty, Choquette-Choo, Sinopalnikov, Weinberger, Vijaykumar, Rogozińska, Herbison, Bandy, Wang, Noland, Moreira, Senter, Eltyshev, Visin, Rasskin, Wei, Cameron, Martins, Hashemi, Klimczak-Plucińska, Batra, Dhand, Nardini, Mein, Zhou, Svensson, Stanway, Chan, Zhou, Carrasqueira, Iljazi, Becker, Fernandez, van Amersfoort, Gordon, Lipschultz, Newlan, yeong Ji, Mohamed, Badola, Black, Millican, McDonell, Nguyen, Sodhia, Greene, Sjoesund, Usui, Sifre, Heuermann, Lago, McNealus, Soares, Kilpatrick, Dixon, Martins, Reid, Singh, Iverson, Görner, Velloso, Wirth, Davidow, Miller, Rahtz, Watson, Risdal,
  Kazemi, Moynihan, Zhang, Kahng, Park, Rahman, Khatwani, Dao, Bardoliwalla, Devanathan, Dumai, Chauhan, Wahltinez, Botarda, Barnes, Barham, Michel, Jin, Georgiev, Culliton, Kuppala, Comanescu, Merhej, Jana, Rokni, Agarwal, Mullins, Saadat, Carthy, Cogan, Perrin, Arnold, Krause, Dai, Garg, Sheth, Ronstrom, Chan, Jordan, Yu, Eccles, Hennigan, Kocisky, Doshi, Jain, Yadav, Meshram, Dharmadhikari, Barkley, Wei, Ye, Han, Kwon, Xu, Shen, Gong, Wei, Cotruta, Kirk, Rao, Giang, Peran, Warkentin, Collins, Barral, Ghahramani, Hadsell, Sculley, Banks, Dragan, Petrov, Vinyals, Dean, Hassabis, Kavukcuoglu, Farabet, Buchatskaya, Borgeaud, Fiedel, Joulin, Kenealy, Dadashi, and Andreev}]{gemmateam2024gemma2improvingopen}
Gemma Team, Morgane Riviere, Shreya Pathak, Pier~Giuseppe Sessa, Cassidy Hardin, Surya Bhupatiraju, Léonard Hussenot, Thomas Mesnard, Bobak Shahriari, Alexandre Ramé, Johan Ferret, Peter Liu, Pouya Tafti, Abe Friesen, Michelle Casbon, Sabela Ramos, Ravin Kumar, Charline~Le Lan, Sammy Jerome, Anton Tsitsulin, Nino Vieillard, Piotr Stanczyk, Sertan Girgin, Nikola Momchev, Matt Hoffman, Shantanu Thakoor, Jean-Bastien Grill, Behnam Neyshabur, Olivier Bachem, Alanna Walton, Aliaksei Severyn, Alicia Parrish, Aliya Ahmad, Allen Hutchison, Alvin Abdagic, Amanda Carl, Amy Shen, Andy Brock, Andy Coenen, Anthony Laforge, Antonia Paterson, Ben Bastian, Bilal Piot, Bo~Wu, Brandon Royal, Charlie Chen, Chintu Kumar, Chris Perry, Chris Welty, Christopher~A. Choquette-Choo, Danila Sinopalnikov, David Weinberger, Dimple Vijaykumar, Dominika Rogozińska, Dustin Herbison, Elisa Bandy, Emma Wang, Eric Noland, Erica Moreira, Evan Senter, Evgenii Eltyshev, Francesco Visin, Gabriel Rasskin, Gary Wei, Glenn Cameron, Gus Martins,
  Hadi Hashemi, Hanna Klimczak-Plucińska, Harleen Batra, Harsh Dhand, Ivan Nardini, Jacinda Mein, Jack Zhou, James Svensson, Jeff Stanway, Jetha Chan, Jin~Peng Zhou, Joana Carrasqueira, Joana Iljazi, Jocelyn Becker, Joe Fernandez, Joost van Amersfoort, Josh Gordon, Josh Lipschultz, Josh Newlan, Ju~yeong Ji, Kareem Mohamed, Kartikeya Badola, Kat Black, Katie Millican, Keelin McDonell, Kelvin Nguyen, Kiranbir Sodhia, Kish Greene, Lars~Lowe Sjoesund, Lauren Usui, Laurent Sifre, Lena Heuermann, Leticia Lago, Lilly McNealus, Livio~Baldini Soares, Logan Kilpatrick, Lucas Dixon, Luciano Martins, Machel Reid, Manvinder Singh, Mark Iverson, Martin Görner, Mat Velloso, Mateo Wirth, Matt Davidow, Matt Miller, Matthew Rahtz, Matthew Watson, Meg Risdal, Mehran Kazemi, Michael Moynihan, Ming Zhang, Minsuk Kahng, Minwoo Park, Mofi Rahman, Mohit Khatwani, Natalie Dao, Nenshad Bardoliwalla, Nesh Devanathan, Neta Dumai, Nilay Chauhan, Oscar Wahltinez, Pankil Botarda, Parker Barnes, Paul Barham, Paul Michel, Pengchong Jin,
  Petko Georgiev, Phil Culliton, Pradeep Kuppala, Ramona Comanescu, Ramona Merhej, Reena Jana, Reza~Ardeshir Rokni, Rishabh Agarwal, Ryan Mullins, Samaneh Saadat, Sara~Mc Carthy, Sarah Cogan, Sarah Perrin, Sébastien M.~R. Arnold, Sebastian Krause, Shengyang Dai, Shruti Garg, Shruti Sheth, Sue Ronstrom, Susan Chan, Timothy Jordan, Ting Yu, Tom Eccles, Tom Hennigan, Tomas Kocisky, Tulsee Doshi, Vihan Jain, Vikas Yadav, Vilobh Meshram, Vishal Dharmadhikari, Warren Barkley, Wei Wei, Wenming Ye, Woohyun Han, Woosuk Kwon, Xiang Xu, Zhe Shen, Zhitao Gong, Zichuan Wei, Victor Cotruta, Phoebe Kirk, Anand Rao, Minh Giang, Ludovic Peran, Tris Warkentin, Eli Collins, Joelle Barral, Zoubin Ghahramani, Raia Hadsell, D.~Sculley, Jeanine Banks, Anca Dragan, Slav Petrov, Oriol Vinyals, Jeff Dean, Demis Hassabis, Koray Kavukcuoglu, Clement Farabet, Elena Buchatskaya, Sebastian Borgeaud, Noah Fiedel, Armand Joulin, Kathleen Kenealy, Robert Dadashi, and Alek Andreev. 2024.
\newblock \href {https://arxiv.org/abs/2408.00118} {Gemma 2: Improving open language models at a practical size}.
\newblock \emph{Preprint}, arXiv:2408.00118.

\bibitem[{Willard and Louf(2023)}]{willard2023efficientguidedgenerationlarge}
Brandon~T. Willard and Rémi Louf. 2023.
\newblock \href {https://arxiv.org/abs/2307.09702} {Efficient guided generation for large language models}.
\newblock \emph{Preprint}, arXiv:2307.09702.

\end{thebibliography}
% \bibliography{acl}

\clearpage
\appendix

% ============
% Dataset Construction
% ============
\section{Dataset Construction} \label{app:dataset-construction}

\begin{figure*}[t]
  \includegraphics[width=\textwidth]{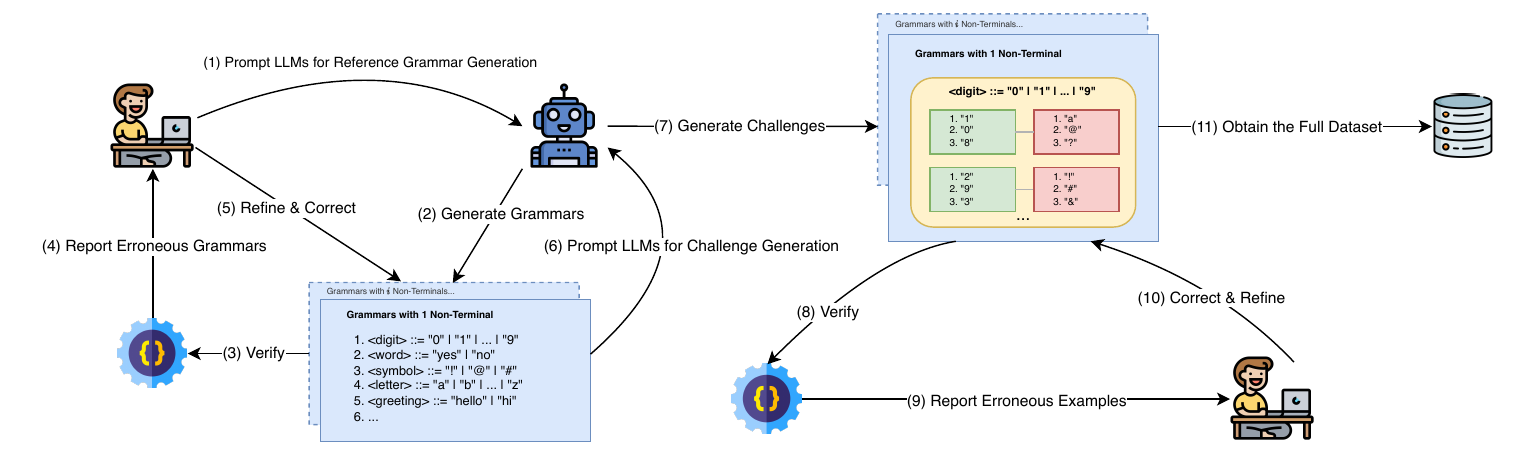}
  \caption{The Dataset Construction Process: (1) GPT-4o is prompted with Prompt Template~\ref{prompt:dc_generate_bnfs} to generate a set of reference grammars; (2) A set of reference grammars $\mathfrak{G^{ref}}=\bigcup^9_{k=1} \mathbb{G}^{ref}_{k}$ are generated by LLMs; (3) A BNF parser is used to check the correctness of each generated reference grammar; (4) Erroneous reference grammars are reported to humans; (5) Reported reference grammars are modified and corrected manually; (6) GPT-4o is prompted with Prompt Template~\ref{prompt:dc_generate_pexamples} and Prompt Template~\ref{prompt:dc_generate_nexamples} to generate challenges for each reference grammars; (7) Challenges are generated by LLMs, in which each challenge consists of 3 positive and 3 negative examples; (8) A BNF parser is used to verify whether positive and negative examples are accepted and rejected by their corresponding reference grammar respectively; (9) Erroneous challenges are reported to humans; (10) Reported challenges are corrected manually; (11) The final dataset consisting of 540 challenges are obtained.}
  \label{fig:dataset_construction}
\end{figure*}
To evaluate the capacity of LLMs in few-shot grammar generation, we present a dedicated dataset. We explain the details of the construction process in this section.

For clear explanation, let $\mathfrak{G}^{ref} = \bigcup_{k=1}^K \mathbb{G}^{ref}_{k}$ be a set of reference grammars, where $\mathbb{G}^{ref}_{k}$ is a set of reference grammars in which each reference grammar $G^{ref}_k$ having exactly $k$ non-terminals and thus its $|R| = k$. Each set of reference grammars $\mathbb{G}^{ref}_{k}=\{G^{ref}_{k,1}, G^{ref}_{k,2}, \dots, G^{ref}_{k,n}\}$ contains $n$ reference grammar with $k$ non-terminals. 

Initially, we constructed $\mathfrak{G}^{ref} = \bigcup_{k=1}^9 \mathbb{G}^{ref}_{k}$. For each number $k$, we prompted GPT-4o to produce $n=10$ reference grammars to yield $\mathbb{G}^{ref}_{k}=\{G^{ref}_{k,1}, G^{ref}_{k,2}, \dots, G^{ref}_{k,10}\}$, with the prompt template demonstrated in Prompt Template~\ref{prompt:dc_generate_bnfs}. In Prompt Template~\ref{prompt:dc_generate_bnfs}, $k$ means the placeholder of the number of non-terminals, and $n$ means the number of reference grammars needed to be produced. However, GPT-4o failed to consistently generate reference grammars in the correct syntax or with the correct number of non-terminals, especially as the number of non-terminals increased. To ensure that the generated reference grammars are syntactically correct and have the correct number of non-terminals, we used a BNF parser to do verification. It takes a $G^{ref}$ and checks whether $valid(G^{ref})$ is true and whether it has the required number of non-terminals. Any reference grammar that is not valid or has a wrong number of non-terminals were manually corrected. For duplicated reference grammars, we prompted GPT-4o to generate an alternative. This resulted in 90 reference grammars (i.e., $|\mathfrak{G}^{ref}| = |\bigcup_{k=1}^{9} \mathbb{G}^{ref}_{k}| = 90$), which are the reference grammars used to generate positive and negative examples subsequently.

For each reference grammar $G^{ref} \in \mathfrak{G}^{ref}$, we prompted GPT-4o to generate $6$ various challenges. For each challenge, GPT-4o is prompted by Prompt Template~\ref{prompt:dc_generate_pexamples} and Prompt Template~\ref{prompt:dc_generate_nexamples} to produce a set of $3$ positive examples ($\mathcal{P} \subseteq \mathcal{L}(G^{ref})$), and a set of three negative examples ($\mathcal{N} \cap \mathcal{L}(G^{ref}) = \emptyset$), respectively. In both Prompt Template~\ref{prompt:dc_generate_pexamples} and Prompt Template~\ref{prompt:dc_generate_nexamples}, $m$ means the number of examples needed to be generated, and \textit{reference\_grammar} means the given reference grammar by which the generated examples should be accepted or rejected. However, we observed that GPT-4o frequently failed to produce valid positive and negative examples, leading to either $\mathcal{P} \not\subseteq \mathcal{L}_(G^{ref})$ or $\mathcal{N} \cap \mathcal{L}_(G^{ref}) \neq \emptyset$, or both. The number of failures tends to increase as the number of non-terminals of $G^{ref}$ increases. To ensure the correctness of the generated challenges, we used a BNF parser to verify whether all given positive examples and negative examples can be accepted and rejected, respectively, by their corresponding $G^{ref}$. Erroneous positive and negative examples were manually corrected. Ultimately, we obtained a dataset consisting of a total of 540 challenges.

We visually summarize the dataset construction procedure in Figure~\ref{fig:dataset_construction}.

% ============
% Metrics
% ============
\section{Grammar Quality Metrics}
\label{appendix:grammar_quality_metrics}
This section covers the formal definitions of the grammar quality metrics.

First define $\Pi_\mathcal{P} \subseteq \Pi$ to be the set of production rules that are used in the left-most derivations of all positive examples in $\mathcal{P}$. That is, the set of rules in $\Pi$ which occur in a sequence of rules $S \rightarrow \alpha_1 \rightarrow \ldots \rightarrow \alpha_n \rightarrow p $ where $p \in \mathcal{P}$, and all rules expand the left-most non-terminal in $\alpha_1, \ldots, \alpha_n$.

% First define a $\mathrm{ProdUsed}$ function for measuring the number of production rules used during derivation for a grammar $\mathcal{G}$: 
% \[
% \mathrm{ProdUsed}(\mathcal{E}^{\mathcal{G}}, \mathcal{P})
% \;=\;
% \Bigl|\{\,a \in R
% \;\bigm|\;
% S \overset{*}{\underset{a}{\Rightarrow}} p
% \}\Bigr|.
% \]
% where $p \in \mathcal{P}$ , and $S$ derives $p$ in zero or more steps using production rule $a \in R$. 

Let $\Pi^*$ be the production rules in $\guess$ and $\Pi^{ref}$ be the production rules in $\bnfref$. Let us define $\diff(\bnfref,\guess) =  |\Pi^{ref}_\mathcal{P}| - |\Pi^*_\mathcal{P}|.$
The average difference in production rules used over the whole set of $k$ solved challenges, $C'$, is given by: 
$$
\diff^\diamond = \frac{1}{k} \sum_{i=1}^k \diff(\bnfref_i,\guess_i).
$$

% We define the difference of the number of used production rules between two grammars as:
% \[
% \mathrm{PUDiff}(\mathcal{E}^{\mathcal{G}_{(1)}},\mathcal{E}^{{\mathcal{G}}_{(2)}},\mathcal{P}) = \mathrm{PU_{(1)}} - \mathrm{PU_{(2)}}
% \]
% where $\mathrm{PU_{(1)}} = \mathrm{ProdUsed}(\mathcal{E}^{\mathcal{G}_{(1)}},\mathcal{P})$ and $\mathrm{PU_{\mathcal{G}^2}} = \mathrm{ProdUsed}(\mathcal{E}^{\mathcal{G}_{(2)}},\mathcal{P})$ to indicate the difference of the number of used production rules between $\mathcal{G}_{(1)}$ and $\mathcal{G}_{(2)}$.

% The $\diff_{avg}$ is given as:
% \begin{align*}
%     \mathrm{\diff_{avg}}(C) = \frac{1}{N} \sum_{i=1}^N \mathrm{PUDiff}(\mathcal{E}^{\mathcal{G}_{ref}}_i,\mathcal{E}^{\mathcal{G}_{llm}}_i,\mathcal{P}_i).
% \end{align*}
% to measure the average difference of used production rules between reference and generated grammars.

% Let $\mathrm{PUDiff}^{(1)}_{(2)}$ be $\mathrm{PUDiff}(\mathcal{E}^{\mathcal{G}_{(1)}},\mathcal{E}^{\mathcal{G}_{(2)}},\mathcal{P})$ and $\mathrm{ProdUsed^{(1)}}$ be $\mathrm{ProdUsed(\mathcal{E}^{\mathcal{G}_{(1)}},\mathcal{P})}$. 

\noindent We define two indicator functions, which indicate when a grammar uses substantially fewer rules than the reference grammar, and substantially more rules than the reference grammar:
% \begin{align*}
% \mathbb{I}_{OF}(\mathcal{E}^{\mathcal{G}_{(1)}},\mathcal{E}^{\mathcal{G}_{(2)}},\mathcal{P}) &= 
% \begin{cases}
% 1 & \text{if } (\mathrm{PUDiff}^{(1)}_{(2)} \\ & \quad > \frac{\mathrm{ProdUsed^{(1)}}}{2}) \\
% 0 & \text{otherwise.}
% \end{cases}
% \end{align*}
\begin{align*}
\mathbb{I}_{OF}(\bnfref, \guess,\mathcal{P}) &= 
\begin{cases}
1 & \text{if } |\Pi^{ref}_\mathcal{P}| - |\Pi^*_\mathcal{P}|
\\ & \quad
> \frac{|\Pi^{ref}_\mathcal{P}|}{2} \\
0 & \text{otherwise.}
\end{cases}
\end{align*}
\begin{align*}
\mathbb{I}_{OG}(\bnfref, \guess,\mathcal{P}) &= 
\begin{cases}
1 & \text{if } |\Pi^{ref}_\mathcal{P}| - |\Pi^*_\mathcal{P}|
\\ & \quad
< -\frac{|\Pi^{ref}_\mathcal{P}|}{2} \\
0 & \text{otherwise.}
\end{cases}
\end{align*}

The metric to estimate whether the generated grammars overfit the examples, is then given by
$$
 \of(C') = \frac{1}{k}\sum_{i=1}^{k} \mathbb{I}_{OF}(\bnfref_i,\guess_i, \mathcal{P}_i).
$$

% We also define an indicator function:
% \begin{align*}
% \mathbb{I}_{OG}(\mathcal{E}^{\mathcal{G}_{(1)}},\mathcal{E}^{\mathcal{G}_{(2)}},\mathcal{P}) &= 
% \begin{cases}
% 1 & \text{if } (\mathrm{PUDiff}^{(1)}_{(2)} \\ &\quad < -\frac{\mathrm{ProdUsed^{(1)}}}{2}) \\
% 0 & \text{otherwise.}
% \end{cases}
% \end{align*}
The metric to estimate whether the generated grammars overgeneralize the examples, $\og$, is given as:
\begin{align*}
    \og(C') = \frac{1}{k}\sum_{i=1}^{k}  \mathbb{I}_{OG}(\bnfref_i, \guess_i, \mathcal{P}_i).
\end{align*}

In addition, the $\tu$ metric, for a given challenge, to measure the percentage of $|\Pi^*|$ taken up by $|\Pi^*_\mathcal{P}|$, indicating the utility of $\guess$, is given as:
$$\tu(G^*,\mathcal{P}) = \frac{|\Pi^*_\mathcal{P}|}{|\Pi^*|}$$ for which lower $TU$ indicates a bunch of irrelevant or nonsensical production rules of $\guess$ while higher $TU$ indicates the opposite. The average utility over $C'$ is given by:$$\tu^\diamond = \frac{1}{k}\sum_{i=1}^{k} \tu(G^*_i,\mathcal{P})$$

% ============
% Direct Prompting
% ============
\section{Direct Prompting}
\label{appendix:DP}

For the DP approach, Prompt Template~\ref{prompt:direct_prompting} is used to prompt LLMs to directly generate a grammar with a given set of positive and negative examples. In Prompt Template~\ref{prompt:direct_prompting}, \textit{positive\_examples} and \textit{negative\_examples} are placeholders for a set of positive and negative examples. In addition, it also specifies a list of requirements LLMs should take care of and obey when generating grammars. 

% ============
% Optimization of BNF Parser for Providing LLM-Friendly Feedback
% ============
\section{Optimization of BNF Parser for Providing LLM-Friendly Feedback}
\label{appendix:OPF}

For the OPF approach, we use the same prompt template used in DP, as shown in Prompt Template~\ref{prompt:direct_prompting}, to prompt LLMs to generate an initial grammar. Then, Prompt Template~\ref{prompt:feedback_in_opf} is used in the iterations of the feedback loop to construct prompts from feedback offered by the BNF parser to LLMs. In Prompt Template~\ref{prompt:feedback_in_opf}, \textit{positive\_examples} and \textit{negative\_examples} are the placeholders for a set of positive and negative examples, \textit{bnf\_grammar} means the previously generated erroneous grammar, and \textit{parser\_feedback} is the placeholder for feedback provided from the BNF parser.

Furthermore, for each feedback given by the BNF parser, in addition to giving essential feedback such as notifying the line number for the place the error occurs, we optimize the BNF parser to also provide LLM-friendly feedback to LLMs, such as the possible reasons for the error or ways to fix it. We have shown some of them in Parser Feedback~\ref{feedback:invalid_rule} and Parser Feedback~\ref{feedback:lack_of_alts}.

In addition, OPF includes a parameter called \textit{max\_turns}, which specifies the maximum number of feedback iterations. If an LLM can generate a valid grammar based on earlier feedback, the algorithm stops early; otherwise, it continues until reaching the specified maximum.

Moreover, it is worth noting that this approach does not aim to follow every component of Reflexion or Self-Refine strictly. For instance, it does not maintain a long-term context or external memory. Instead, it uses only the most recent feedback in each turn to guide self-refinement. Concretely, in each feedback iteration, an LLM is provided with the previously generated erroneous grammar, corresponding positive and negative examples, and the latest feedback, to produce revised grammars. Therefore, since, in each iteration, an LLM may produce similar and even the same grammar as the previous ones, especially when they fail to fix the previous errors leading to the same feedback provided by the parser, we set the temperature to 0.3 to expect to enable LLMs to generate more diverse grammars even for encountering the same feedback, to optimize the performance. This approach highlights the optimization from the perspective of the parser to provide more LLM-friendly feedback. However, due to limited space and the trivial-yet-complex optimization process, for the details of the optimization of the BNF parser, please refer to our source code and the comments.

% We have also demonstrated the pseudocode of OPF in Algorithm~\ref{algo:opf}, in which $\mathcal{P}$ and  $\mathcal{N}$ mean a set of positive and negative examples respectively, $max\_turns$ means the maximum turns for the feedback loop, and $LLM$ is an LLM to generate grammars. 

% \begin{algorithm*}[t]
% \caption{OPF}
% \label{algo:opf}
% \begin{algorithmic}[1]
% \Procedure{GenerateGrammar}{$\mathcal{P}, \mathcal{N}, max\_turns, LLM$}
%     \State $prompt \gets \textsc{PromptGenerator}$ \Comment{Use prompt template from Prompt~\ref{prompt:direct_prompting}}
%     \State $\mathcal{G} \gets LLM(prompt)$ \Comment{Use LLM to get an initial grammar}
%     \For{$i \gets 1$ \textbf{to} \textit{max\_turns}}
%         \State $parsable, error \gets \textsc{Parse}(\mathcal{G})$ \Comment{Parse the given grammar using the BNF parser}
%         \If{not $parsable$}
%             \State $feedback \gets \textsc{FeedBackPromptGenerator}$ \Comment{Use prompt template from Prompt~\ref{prompt:feedback_in_opf}} 
%             \State $\mathcal{G} \gets LLM(feedback)$ \Comment{Use LLM to get a new grammar}
%         \EndIf
%     \EndFor
%     \Return $\mathcal{G}$
% \EndProcedure
% \end{algorithmic}
% \end{algorithm*}

% ============
% LLM-Driven Heuristic Mutation Genetic Algorithm
% ============
\section{LLM-Driven Hybrid Genetic Algorithm}
\label{appendix:LDHGA}

The pseudocode of \NAME is presented in Algorithm~\ref{algo:ldhga}, with detailed descriptions of its primary functions including $\mathrm{Fitness}$, $\mathrm{Select}$, $\mathrm{Cross}$, and $\mathrm{Mutate}$ provided in Section~\ref{sec:ldhm_ga}. In Algorithm~\ref{algo:ldhga}, we note $\mathrm{Fitness}$ function as \textsc{Fitness}, $\mathrm{Select}$ function as \textsc{Select}, $\mathrm{Cross}$ function as \textsc{Cross}, and $\mathrm{Mutate}$ function as \textsc{Mutate}.

As shown in Algorithm~\ref{algo:ldhga}, it takes seven parameters: $\mathcal{P}$ and $\mathcal{N}$ means a set of positive and negative examples respectively, $k$ indicates \emph{population size}, $g$ represents \emph{generations} which means the number of iterations of evolution, $\rho$ means the \emph{crossover rate}, $\mu$ means the \emph{mutation rate}, and $LLM$ means an LLM which takes a prompt and returns a response. In addition, \textsc{PromptGenerator} means to generate a prompt from the prompt template shown in Prompt Template~\ref{prompt:direct_prompting}. \textsc{MaxFitnessScore} is a constant indicating the highest fitness score any candidate grammar can achieve and due to each challenge in the constructed dataset only having 3 positive and 3 negative examples, the highest fitness score is 6. We thus set \textsc{MaxFitnessScore} to 6.

In addition, for \textit{LLMMut}, we have shown the prompt template used to prompt LLMs to mutate a given grammar in Prompt Template~\ref{prompt:llm_driven_mutation}, in which \textit{bnf\_grammar} means the placeholder for a candidate grammar while \textit{positive\_examples} and \textit{negative\_examples} means a set of positive and negative examples respectively.

\begin{algorithm*}[h]
\caption{\NAME}
\label{algo:ldhga}
\begin{algorithmic}[1]
\Procedure{GenerateGrammar}{$\mathcal{P}, \mathcal{N}, k, g, \rho, \mu, LLM$}
    \State $population \gets []$ \Comment{Initialize population as an empty list}
    \State $G^*.best \gets \texttt{null}$ \Comment{Keep track of overall best grammar}
    \State $fitness.best \gets -1$ \Comment{Track highest fitness found so far}
    \For{$i \gets 1$ \textbf{to} $k$}
        \State $prompt \gets \textsc{PromptGenerator}$ \Comment{Use prompt template from Prompt Template~\ref{prompt:direct_prompting}}
        \State $G^* \gets LLM(prompt)$ \Comment{Use LLM to get an initial candidate}
        \State $score \gets \textsc{Fitness}(G^*, \mathcal{P}, \mathcal{N})$ \Comment{Compute fitness}
        \If{$score = \textsc{MaxFitnessScore}$} 
            \State \Return $G^*$ \Comment{Return early if perfect score is achieved}
        \EndIf
        \State $population \gets population \parallel [G^*]$ \Comment{Add candidate to population}
    \EndFor
    \For{$i \gets 1$ \textbf{to} $g$}
        \State $fitnessScores \gets []$ 
        \For{$G^* \in population$}
            \State $score \gets \textsc{Fitness}(G^*, \mathcal{P}, \mathcal{N})$
            \State $fitnessScores \gets fitnessScores \parallel [(score, G^*)]$ \Comment{Add a tuple of score and grammar}
            \If{$score > fitness.best$}
                \State $fitness.best \gets score$
                \State $G^*.best \gets G^*$
            \EndIf
        \EndFor
        \If{$fitness.best = \textsc{MaxFitnessScore}$}
                \State \Return $G^*.best$ \Comment{Return the best grammar found} 
        \EndIf
        % \State \textsc{Sort}($fitnessScores$) \Comment{Sort by score in descending order}

        \State $\mathbb{G} = [G^* \mid (score, G^*) \in fitnessScores]$
        \State $S = [score \mid (score,G^*) \in fitnessScores]$
        \State $selected \gets \textsc{Select}(\mathbb{G},S)$
        \State $population.new \gets []$

        \While{$|population.new| < k$}
            \State $ G^*_a,G^*_b \gets \textsc{RandomChoice}(selected)$
            \State $G^* \gets \textsc{cross}(G^*_a,G^*_b, \rho)$
            \If{$\textsc{Uniform}(0,1) < \mu$}
                \State $G^* \gets \textsc{mutate}(G^* ,\mathcal{P}, \mathcal{N}, LLM)$
            \EndIf
            \State $population.new \gets population.new \parallel [G^*]$
        \EndWhile
        
        \State $population \gets population.new$ \Comment{Proceed to the next generation}
    \EndFor
    \Return $G^*.best$
\EndProcedure
\end{algorithmic}
\end{algorithm*}

\section{Additional Results}
\label{appendix:additional_results}
In addition to the results shown in Table~\ref{tab:results_for_all}, we have shown 7 more results and discussed them respectively in subsections~\ref{appendix:nonterminals_results},~\ref{appendix:prs_results},~\ref{appendix:fgi_results},~\ref{appendix:tu_results}, and~\ref{appendix:multiple_experiments}.

\subsection{Results for $C_1$, $C_2$, and $C_3$}
\label{appendix:nonterminals_results}
Table~\ref{tab:nonterminals_results} presents the results categorized into three subsets: $C_1$, $C_2$, and $C_3$. The subset $C_1$ includes challenges where the reference grammars have $1\sim3$ non-terminals, $C_2$ are those with $4\sim6$ non-terminals, and $C_3$ consists of challenges with $7\sim9$ non-terminals. Therefore, it aims to demonstrate and analyze the performance of LLMs as the number of non-terminals increases.

As the results are shown in Table~\ref{tab:nonterminals_results}, as the number of non-terminals increases, both $\sx$ and $\se$ decrease across all LLMs. While DP and OPF exhibit suboptimal and unsatisfactory performance, \NAME consistently demonstrates and contributes substantial improvements across most LLMs, even as the number of non-terminals increases. For example, in the case of \textit{GPT-4o}, \NAME increases the $\se$ by 21\% compared to DP and OPF on $C_3$. Similarly, with \textit{GPT-3.5-Turbo}, compared to DP and OPF, \NAME improves the $\se$ by 30\% and 28\% on $C_2$ and 20\% and 19\% on $C_3$, respectively.

\subsection{Results for $P_1$, $P_2$, and $P_3$}
\label{appendix:prs_results}
Table~\ref{tab:prs_results} demonstrates the results grouped into three subsets: $P_1$, $P_2$, and $P_3$. The subset $P_1$ consists of challenges where the reference grammars have $1\sim6$ production rules, $P_2$ includes those with $7\sim15$ production rules, and $P_3$ consists of challenges with the number of production rules greater than $16$. Therefore, it aims to show and analyze the performance of LLMs as the number of production rules increases. 

Similar to the results of $C_1$, $C_2$, and $C_3$, as the number of production rules increases, both $\sx$ and $\se$ decrease across all LLMs. Nevertheless, \NAME can still steadily improve both $\se$ and $\sx$, even as the number of production rules increases.

\subsection{Results for $\diff$, $\of$, and $\og$ Metrics}
\label{appendix:fgi_results}
To investigate whether LLMs generate grammars in an overfitted manner and whether \NAME improves performance through overfitting, as well as to examine whether LLMs and our method produce overly generalized grammars, we employed the three evaluation metrics: $\diff$, $\of$, $\og$. The results are presented in Table~\ref{tab:fgi_diff}, ~\ref{tab:fgi_of}, and~\ref{tab:fgi_og}, respectively. The notation “N/A” indicates inapplicability. Since these three metrics are only applicable when a grammar possesses correct semantics, “N/A” thus signifies that no grammar in the evaluation set exhibits correct semantics.

Through the $\diff$ metric, as shown in Tabel~\ref{tab:fgi_diff}, we observe that the number of production rules used in derivations of the generated grammars and the reference grammars does not differ significantly on average. However, as the number of production rules in the reference grammar increases, the $\diff^\diamond$ exhibits a slight upward trend. The $\diff^\diamond$ is almost always positive across most LLMs and methods, which indicates that, in most cases, the number of production rules used by the generated grammars is lower than that of the reference grammars.

Furthermore, Table~\ref{tab:fgi_of} presents the $\of$ metric. For some models, particularly \textit{GPT-3.5-Turbo}, an increasing number of production rules corresponds to a certain degree of overfitting. Nevertheless, on average, most LLMs do not exhibit significant overfitting. Additionally, we observe that in \NAME, the $\of$ metric does not show a significant difference compared to the baselines, indicating that \NAME does not improve performance through overfitting.

Additionally, we present the $\og$ metric in Table~\ref{tab:fgi_og}. We observed that, for some models, such as \textit{Qwen:72b-Instruct}, as the number of production rules in the reference grammar increases, they tend to generate overly generalized grammars. Nevertheless, on average, most LLMs do not tend to generate overgeneralized grammars.

\subsection{Results for $\tu$ Metrics}
\label{appendix:tu_results}
To investigate whether LLMs generate irrelevant production rules, we employ the $\tu$ metric. For example, given the challenge shown in Figure~\ref{fig:bnf_generation}, a generated grammar might be:

\vspace{1em}
\begin{minipage}{\linewidth}
\begin{verbatim}
<stmt> ::= <func> "(" <args> ")"
<args> ::= <expr> | <expr> "," <args>
<expr> ::= <char> | <number>
<func> ::= <char> <func> | <char>
<char> ::= "a" | ... | "z"
<number> ::= "0" | ... | "9"
<hello> ::= "hello"
<world> ::= "world"

\end{verbatim}
\end{minipage}
\vspace{1em}
in which:

\vspace{1em}
\begin{minipage}{\linewidth}
\begin{verbatim}
<hello> ::= "hello"
<world> ::= "world"

\end{verbatim}
\end{minipage} 
\vspace{1em}
are two irrelevant production rules.

As shown in Table~\ref{tab:fgi_tu}, on average, across both baselines and in \NAME, TU remains relatively high, indicating that LLMs do not tend to produce irrelevant production rules. However, as the number of production rules increases, $\tu$ shows a tendency of declination. Nevertheless, this does not imply that LLMs generate more irrelevant rules. Considering the results from $\og$, we think this decrease may more likely be attributable to the generated grammar becoming more generalized.

\subsection{Results for Robustness Evaluation}
\label{appendix:multiple_experiments}

Since \NAME requires setting the temperature greater than 0 which we set to 0.7, we repeated 5 independent experiments for both \textit{GPT-4o} and \textit{GPT-3.5-Turbo} to ensure the temperature does not affect the results significantly.

The results are demonstrated in Table~\ref{tab:5_times}, in which each row means the results of one experiment. The averages of syntax correctness of \textit{GPT-4o} and \textit{GPT-3.5-Turbo} are 95.8\% and 98.6\% and the standard deviations are 0.4\% and 0.49\%, respectively. The averages of semantic correctness of \textit{GPT-4o} and \textit{GPT-3.5-Turbo} are 93.2\% and 61.6\% and the standard deviations are 0.4\% and 0.49\%. Therefore, it indicates that although we set the temperature to 0.7 in \NAME, the fluctuation of both syntax correctness and semantic correctness are very slight and the performance across multiple experiments stays steady. Thus, the results demonstrated the robustness of our proposed method, \NAME.

% ===============
% Move prompts here for a better layout
% ===============

\begin{promptblock}{Generate a Grammar Directly with a Given Set Positive and Negative Examples}
\label{prompt:direct_prompting}
Given a set of positive and negative examples, generate the Backus–Naur Form (BNF) grammar that accepts all positive examples and rejects all negative examples.\\
1. Only generate the standard BNF grammar;\\
2. The generated BNF grammar MUST accept all positive examples and reject all negative examples;\\
3. Each terminal symbol MUST be quoted with double quotes and MUST NOT escape double quotes or pipeline in terminal symbols;\\
4. Pay special attention to whether spaces, line breaks, or other special symbols are required between each symbol, and if so, these need to be explicitly specified, e.g. <term> ::= "1" "+" "2" can handle "1+2" but not "1 + 2" while <term> ::= "1" " " "+" " " "2" can handle "1 + 2" but not "1+2";\\
5. The entry point of the generated BNF grammar MUST be the non-terminal symbol in the first production rule;\\
6. Only the generated BNF should be wrapped in a pair of triple backtick;\\
7. Do NOT output any additional texts, comments, or explanations.\\
\vspace{\baselineskip}
===Positive Examples===\\
\{\textit{positive\_examples}\}\\
===Negative Examples===\\
\{\textit{negative\_examples}\}\\
\end{promptblock}

% \vspace{1.5em}

% Syntax and semantic correctness results of gpt-4o and gpt-3.5-turbo with 5 times
\begin{table}
\centering
\begin{tabularx}{\linewidth}{X X X}
\toprule
\textbf{Experiment} & \textbf{$\sx$} & \textbf{$\se$} \\
\midrule
\multicolumn{3}{c}{\small\textbf{GPT-4o}} \\
\midrule
1st & 95 & 93 \\
2nd & 96 & 93 \\
3rd & 96 & 94 \\
4th & 96 & 93 \\
5th & 96 & 93 \\
\midrule
\multicolumn{3}{c}{\small\textbf{GPT-3.5-Turbo}} \\
\midrule
1st & 98 & 62 \\
2nd & 98 & 62 \\
3rd & 99 & 62 \\
4th & 99 & 61 \\
5th & 99 & 61 \\
\bottomrule
\end{tabularx}
\caption{Results of Syntax and Semantic Correctness for \NAME with \textit{GPT-4o} and \textit{GPT-3.5-Turbo} on Grammar Generation by Conducting 5 Independent Experiments (\%)}
\label{tab:5_times}
\end{table}

\begin{feedbackblock}{Invalid Production Rule}
\label{feedback:invalid_rule}
This error is likely due to not satisfying one of the following requirements:\\
1. A rule MUST start with a non-terminal definition;\\
2. A non-terminal symbol MUST be in angle brackets, e.g. <non-terminal>;\\
3. A non-terminal definition must be followed by '::=' to indicate the start of the right-hand side;\\
\end{feedbackblock}

\begin{promptblock}{LLM-Driven Mutation}
\label{prompt:llm_driven_mutation}
Modify the following BNF grammar slightly to improve its acceptance of the positive examples and rejection of the negative examples.\\
\vspace{\baselineskip}
===BNF Grammar===\\
\{\textit{bnf\_grammar}\}\\
\vspace{\baselineskip}
===Positive Examples===\\
\{\textit{positive\_examples}\}\\
===Negative Examples===\\
\{\textit{negative\_examples}\}\\
\vspace{\baselineskip}
Only output the modified BNF grammar wrapped in triple backticks.
\end{promptblock}

\begin{promptblock}[t]{Generate Grammars}
\label{prompt:dc_generate_bnfs}
Generate a list of random standard Backus-Naur Form (BNF) grammar with the following constraints:\\
1. Each generated BNF grammar MUST be SELF-CONTAINED and VALID, which means it should be able to recognize a valid string;\\
2. Each generated BNF grammar MUST have exactly $\{k\}$ lines;\\
3. Each generated BNF grammar MUST be unique;\\
4. Each generated BNF grammar MUST be separated by a newline in addition to the linebreak;\\
5. For each generated BNF grammar, the entry point MUST be at the first line;\\
6. Only generate $\{n\}$ BNF grammars;\\
7. Only output BNF grammars WITHOUT any additional text or code block, like "\verb|```|".
\end{promptblock}

\begin{promptblock}{Generate Negative Examples with a Given Grammar}
\label{prompt:dc_generate_nexamples}
Generate a list of negative examples with the following constraints:\\
1. Each example MUST be separated by a newline in addition to the linebreak;\\
2. Only output examples WITHOUT any additional text or code block, like "\verb|```|";\\
3. Only output $\{m\}$ examples;\\
4. Each example MUST be generated based on the given BNF grammar;\\
5. Each example should be greatly related to the given BNF grammar, but ensure it is NOT a valid string for the given BNF grammar.\\

For example, given the following BNF grammar:\\
<term> ::=  "0" | "1" | "2"\\
you should output negative examples like:\\
6\\
\vspace{\baselineskip}
*\\
\vspace{\baselineskip}
9\\
\vspace{\baselineskip}
Then, the given BNF grammar is:\\
\{\textit{reference\_grammar}\}
\end{promptblock}

\begin{feedbackblock}{Lack of Alternatives}
\label{feedback:lack_of_alts}
This error is likely due to the reason that the right-hand side is not defined after '::='.
\end{feedbackblock}

\begin{promptblock}{Feedback Prompt in OPF}
\label{prompt:feedback_in_opf}
Given a set of positive and negative examples, generate the Backus–Naur Form (BNF) grammar that accepts all positive examples and rejects all negative examples.\\
1. Only generate the standard BNF grammar;\\
2. The generated BNF grammar MUST accept all positive examples and reject all negative examples;\\
3. Each terminal symbol MUST be quoted with double quotes and MUST NOT escape double quotes or pipeline in terminal symbols;\\
4. Pay special attention to whether spaces, line breaks, or other special symbols are required between each symbol, and if so, these need to be explicitly specified, e.g. <term> ::= "1" "+" "2" can handle "1+2" but not "1 + 2" while <term> ::= "1" " " "+" " " "2" can handle "1 + 2" but not "1+2";\\
5. The entry point of the generated BNF grammar MUST be the non-terminal symbol in the first production rule;\\
6. Only the generated BNF should be wrapped in a pair of triple backtick;\\
7. Do NOT output any additional texts, comments, or explanations.\\
\vspace{\baselineskip}
===Positive Examples===\\
\{\textit{positive\_examples}\}\\
===Negative Examples===\\
\{\textit{negative\_examples}\}\\
\vspace{\baselineskip}
===Generated BNF===\\
\{{\textit{bnf\_grammar}}\}\\
\vspace{\baselineskip}
===Feedback===\\
The generated BNF grammar has incorrect syntax and please consider fixing it by referring to the feedback.\\
Here is the feedback from the BNF parser:\\
\{\textit{parser\_feedback}\}
\end{promptblock}

\begin{promptblock}{Generate Positive Examples with a Given Grammar}
\label{prompt:dc_generate_pexamples}
Generate a list of positive examples with the following constraints:\\
1. Each example MUST be separated by a newline in addition to the linebreak;\\
2. Only output examples WITHOUT any additional text or code block, like "\verb|```|";\\
3. Only output $\{m\}$ examples;\\
4. Each example MUST be generated based on the given BNF grammar;\\
5. Pay attention to whether the whitespaces are allowed between symbols.\\

For example, given the following BNF grammar:\\
<term> ::=  "0" | "1" | "2"\\
you should output positive examples like:\\
0\\
\vspace{\baselineskip}
1\\
\vspace{\baselineskip}
2\\
\vspace{\baselineskip}
Then, the given BNF grammar is:\\
\{\textit{reference\_grammar}\}
\end{promptblock}
\vspace{3.5em}

% Syntax and semantic correctness results of LLMS for compounds grouped
\begin{table*}
\centering
\begin{tabularx}{\textwidth}{l X X X X X X}
\toprule
\textbf{Challenge Set} & \textbf{$SX_{DP}$} & \textbf{$SX_{OPF}$} & \textbf{$SX_{\textbf{\NAME}}$} & \textbf{$SE_{DP}$} & \textbf{$SE_{OPF}$} & \textbf{$SE_{\textbf{\NAME}}$} \\
% gpt-4o
\midrule
\multicolumn{7}{c}{\small\textbf{GPT-4o}} \\
\midrule
$C_1$ 
& 100 
& 100 
& 100 
& 99 
& 99
& 100~\textcolor{blue}{\scriptsize$\uparrow$1}~\textcolor{red}{\scriptsize$\uparrow$1} \\
$C_2$ 
& 100 
& 100 
& 100 
& 93 
& 95~\textcolor{blue}{\scriptsize$\uparrow$2}  
& 100~\textcolor{blue}{\scriptsize$\uparrow$7}~\textcolor{red}{\scriptsize$\uparrow$5} \\
$C_3$ 
& 79 
& 92~\textcolor{blue}{\scriptsize$\uparrow$13}   
& 87~\textcolor{blue}{\scriptsize$\uparrow$8}~\textcolor{red}{\scriptsize$\downarrow$5} 
& 59 
& 59 
& 80~\textcolor{blue}{\scriptsize$\uparrow$21}~\textcolor{red}{\scriptsize$\uparrow$21} \\
All
& 93 
& 97~\textcolor{blue}{\scriptsize$\uparrow$4}   
& 96~\textcolor{blue}{\scriptsize$\uparrow$3}~\textcolor{red}{\scriptsize$\downarrow$1} 
& 84 
& 85~\textcolor{blue}{\scriptsize$\uparrow$1}   
& 93~\textcolor{blue}{\scriptsize$\uparrow$9}~\textcolor{red}{\scriptsize$\uparrow$8} \\
% gpt-4.5-turbo
\midrule
\multicolumn{7}{c}{\small\textbf{GPT-3.5-Turbo}} \\
\midrule
$C_1$ 
& 98 
& 97~\textcolor{blue}{\scriptsize$\downarrow$1}   
& 100~\textcolor{blue}{\scriptsize$\uparrow$2}~\textcolor{red}{\scriptsize$\uparrow$3}
& 72 
& 71~\textcolor{blue}{\scriptsize$\downarrow$1}   
& 93~\textcolor{blue}{\scriptsize$\uparrow$21}~\textcolor{red}{\scriptsize$\uparrow$22} \\
$C_2$ 
& 98 
& 99~\textcolor{blue}{\scriptsize$\uparrow$1}   
& 100~\textcolor{blue}{\scriptsize$\uparrow$2}~\textcolor{red}{\scriptsize$\uparrow$1}
& 28 
& 30~\textcolor{blue}{\scriptsize$\uparrow$2}   
& 58~\textcolor{blue}{\scriptsize$\uparrow$30}~\textcolor{red}{\scriptsize$\uparrow$28} \\
$C_3$ 
& 84 
& 90~\textcolor{blue}{\scriptsize$\uparrow$6}   
& 96~\textcolor{blue}{\scriptsize$\uparrow$12}~\textcolor{red}{\scriptsize$\uparrow$6}
& 11 
& 12~\textcolor{blue}{\scriptsize$\uparrow$1}   
& 31~\textcolor{blue}{\scriptsize$\uparrow$20}~\textcolor{red}{\scriptsize$\uparrow$19} \\
All 
& 94 
& 95~\textcolor{blue}{\scriptsize$\uparrow$1}   
& 99~\textcolor{blue}{\scriptsize$\uparrow$5}~\textcolor{red}{\scriptsize$\uparrow$4}
& 37 
& 38~\textcolor{blue}{\scriptsize$\uparrow$1}   
& 61~\textcolor{blue}{\scriptsize$\uparrow$24}~\textcolor{red}{\scriptsize$\uparrow$23} \\
% qwen:72b-chat
\midrule
\multicolumn{7}{c}{\small\textbf{Qwen:72b-Chat}} \\
\midrule
$C_1$ 
& 73 
& 76~\textcolor{blue}{\scriptsize$\uparrow$3}   
& 96~\textcolor{blue}{\scriptsize$\uparrow$23}~\textcolor{red}{\scriptsize$\uparrow$20}
& 52 
& 53~\textcolor{blue}{\scriptsize$\uparrow$1}   
& 76~\textcolor{blue}{\scriptsize$\uparrow$24}~\textcolor{red}{\scriptsize$\uparrow$23} \\
$C_2$ 
& 48
& 48~\textcolor{blue}   
& 77~\textcolor{blue}{\scriptsize$\uparrow$29}~\textcolor{red}{\scriptsize$\uparrow$29}
& 6 
& 8~\textcolor{blue}{\scriptsize$\uparrow$2}   
& 26~\textcolor{blue}{\scriptsize$\uparrow$20}~\textcolor{red}{\scriptsize$\uparrow$18} \\
$C_3$ 
& 20 
& 23~\textcolor{blue}{\scriptsize$\uparrow$3}   
& 56~\textcolor{blue}{\scriptsize$\uparrow$36}~\textcolor{red}{\scriptsize$\uparrow$33}
& 1 
& 2~\textcolor{blue}{\scriptsize$\uparrow$1}   
& 11~\textcolor{blue}{\scriptsize$\uparrow$10}~\textcolor{red}{\scriptsize$\uparrow$9} \\
All 
& 47 
& 49~\textcolor{blue}{\scriptsize$\uparrow$2}   
& 76~\textcolor{blue}{\scriptsize$\uparrow$29}~\textcolor{red}{\scriptsize$\uparrow$27}
& 20 
& 21~\textcolor{blue}{\scriptsize$\uparrow$1}   
& 38~\textcolor{blue}{\scriptsize$\uparrow$18}~\textcolor{red}{\scriptsize$\uparrow$17} \\
% llama3:70b-instruct
\midrule
\multicolumn{7}{c}{\small\textbf{Llama3:70b-Instruct}} \\
\midrule
$C_1$ 
& 88 
& 90~\textcolor{blue}{\scriptsize$\uparrow$2}   
& 97~\textcolor{blue}{\scriptsize$\uparrow$9}~\textcolor{red}{\scriptsize$\uparrow$7}
& 78 
& 77~\textcolor{blue}{\scriptsize$\downarrow$1}   
& 94~\textcolor{blue}{\scriptsize$\uparrow$16}~\textcolor{red}{\scriptsize$\uparrow$17} \\
$C_2$ 
& 54
& 60~\textcolor{blue}{\scriptsize$\uparrow$6}   
& 76~\textcolor{blue}{\scriptsize$\uparrow$22}~\textcolor{red}{\scriptsize$\uparrow$16}
& 31 
& 35~\textcolor{blue}{\scriptsize$\uparrow$4}   
& 61~\textcolor{blue}{\scriptsize$\uparrow$30}~\textcolor{red}{\scriptsize$\uparrow$26} \\
$C_3$ 
& 28 
& 34~\textcolor{blue}{\scriptsize$\uparrow$6}   
& 52~\textcolor{blue}{\scriptsize$\uparrow$24}~\textcolor{red}{\scriptsize$\uparrow$18}
& 15 
& 14~\textcolor{blue}{\scriptsize$\downarrow$1}   
& 29~\textcolor{blue}{\scriptsize$\uparrow$14}~\textcolor{red}{\scriptsize$\uparrow$15} \\
All 
& 57 
& 61~\textcolor{blue}{\scriptsize$\uparrow$4}   
& 75~\textcolor{blue}{\scriptsize$\uparrow$18}~\textcolor{red}{\scriptsize$\uparrow$14}
& 41 
& 42~\textcolor{blue}{\scriptsize$\uparrow$1}   
& 61~\textcolor{blue}{\scriptsize$\uparrow$20}~\textcolor{red}{\scriptsize$\uparrow$19} \\
% gemma2:27b-instruct
\midrule
\multicolumn{7}{c}{\small\textbf{Gemma2:27b-Instruct}} \\
\midrule
$C_1$ 
& 99 
& 100~\textcolor{blue}{\scriptsize$\uparrow$1}   
& 100~\textcolor{blue}{\scriptsize$\uparrow$1}
& 91 
& 92~\textcolor{blue}{\scriptsize$\uparrow$1}   
& 98~\textcolor{blue}{\scriptsize$\uparrow$7}~\textcolor{red}{\scriptsize$\uparrow$6} \\
$C_2$ 
& 97 
& 97 
& 99~\textcolor{blue}{\scriptsize$\uparrow$2}~\textcolor{red}{\scriptsize$\uparrow$2}
& 49 
& 49 
& 84~\textcolor{blue}{\scriptsize$\uparrow$35}~\textcolor{red}{\scriptsize$\uparrow$35} \\
$C_3$ 
& 76 
& 79~\textcolor{blue}{\scriptsize$\uparrow$3}   
& 93~\textcolor{blue}{\scriptsize$\uparrow$17}~\textcolor{red}{\scriptsize$\uparrow$14}
& 26 
& 29~\textcolor{blue}{\scriptsize$\uparrow$3}   
& 54~\textcolor{blue}{\scriptsize$\uparrow$28}~\textcolor{red}{\scriptsize$\uparrow$25} \\
All 
& 91 
& 92~\textcolor{blue}{\scriptsize$\uparrow$1}   
& 98~\textcolor{blue}{\scriptsize$\uparrow$7}~\textcolor{red}{\scriptsize$\uparrow$6}
& 56 
& 57~\textcolor{blue}{\scriptsize$\uparrow$1}   
& 79~\textcolor{blue}{\scriptsize$\uparrow$23}~\textcolor{red}{\scriptsize$\uparrow$22} \\
% mistral:7b-instruct
\midrule
\multicolumn{7}{c}{\small\textbf{Mistral:7b-Instruct}} \\
\midrule
$C_1$ 
& 1 
& 25~\textcolor{blue}{\scriptsize$\uparrow$24}   
& 3~\textcolor{blue}{\scriptsize$\uparrow$2}~\textcolor{red}{\scriptsize$\downarrow$22}
& 0 
& 17~\textcolor{blue}{\scriptsize$\uparrow$17}   
& 2~\textcolor{blue}{\scriptsize$\uparrow$2}~\textcolor{red}{\scriptsize$\downarrow$15} \\
$C_2$ 
& 1 
& 20~\textcolor{blue}{\scriptsize$\uparrow$19}    
& 1~\textcolor{red}{\scriptsize$\downarrow$19}
& 1 
& 6~\textcolor{blue}{\scriptsize$\uparrow$5}    
& 0~\textcolor{blue}{\scriptsize$\downarrow$1}~\textcolor{red}{\scriptsize$\downarrow$6} \\
$C_3$ 
& 1 
& 11~\textcolor{blue}{\scriptsize$\uparrow$10}   
& 0~\textcolor{blue}{\scriptsize$\downarrow$1}~\textcolor{red}{\scriptsize$\downarrow$11}
& 0 
& 1~\textcolor{blue}{\scriptsize$\uparrow$1}   
& 0~\textcolor{red}{\scriptsize$\downarrow$1}   \\
All 
& 1 
& 19~\textcolor{blue}{\scriptsize$\uparrow$18}   
& 1~\textcolor{red}{\scriptsize$\downarrow$18}
& 0 
& 8~\textcolor{blue}{\scriptsize$\uparrow$8}   
& 1~\textcolor{blue}{\scriptsize$\uparrow$1}~\textcolor{red}{\scriptsize$\downarrow$7} \\
% codestral:22b
\midrule
\multicolumn{7}{c}{\small\textbf{Codestral:22b}} \\
\midrule
$C_1$ 
& 82 
& 96~\textcolor{blue}{\scriptsize$\uparrow$14}   
& 99~\textcolor{blue}{\scriptsize$\uparrow$17}~\textcolor{red}{\scriptsize$\uparrow$3}
& 82 
& 92~\textcolor{blue}{\scriptsize$\uparrow$10}   
& 98~\textcolor{blue}{\scriptsize$\uparrow$16}~\textcolor{red}{\scriptsize$\uparrow$6} \\
$C_2$ 
& 53 
& 77~\textcolor{blue}{\scriptsize$\uparrow$24}   
& 86~\textcolor{blue}{\scriptsize$\uparrow$33}~\textcolor{red}{\scriptsize$\uparrow$9}
& 36 
& 45~\textcolor{blue}{\scriptsize$\uparrow$9}   
& 69~\textcolor{blue}{\scriptsize$\uparrow$33}~\textcolor{red}{\scriptsize$\uparrow$24} \\
$C_3$ 
& 23 
& 39~\textcolor{blue}{\scriptsize$\uparrow$16}   
& 57~\textcolor{blue}{\scriptsize$\uparrow$34}~\textcolor{red}{\scriptsize$\uparrow$18}
& 15 
& 19~\textcolor{blue}{\scriptsize$\uparrow$4}   
& 33~\textcolor{blue}{\scriptsize$\uparrow$18}~\textcolor{red}{\scriptsize$\uparrow$14} \\
All 
& 53 
& 71~\textcolor{blue}{\scriptsize$\uparrow$18}   
& 80~\textcolor{blue}{\scriptsize$\uparrow$27}~\textcolor{red}{\scriptsize$\uparrow$9}
& 44 
& 52~\textcolor{blue}{\scriptsize$\uparrow$8}   
& 67~\textcolor{blue}{\scriptsize$\uparrow$23}~\textcolor{red}{\scriptsize$\uparrow$15} \\
% starcoder2:15b-instruct
\midrule
\multicolumn{7}{c}{\small\textbf{Starcoder2:15b-Instruct}} \\
\midrule
$C_1$ 
& 97 
& 68~\textcolor{blue}{\scriptsize$\downarrow$29}   
& 100~\textcolor{blue}{\scriptsize$\uparrow$3}~\textcolor{red}{\scriptsize$\uparrow$32}
& 67 
& 42~\textcolor{blue}{\scriptsize$\downarrow$25}   
& 84~\textcolor{blue}{\scriptsize$\uparrow$17}~\textcolor{red}{\scriptsize$\uparrow$42} \\
$C_2$ 
& 73 
& 65~\textcolor{blue}{\scriptsize$\downarrow$8}   
& 99~\textcolor{blue}{\scriptsize$\uparrow$26}~\textcolor{red}{\scriptsize$\uparrow$34}
& 14 
& 12~\textcolor{blue}{\scriptsize$\downarrow$2}   
& 31~\textcolor{blue}{\scriptsize$\uparrow$17}~\textcolor{red}{\scriptsize$\uparrow$19} \\
$C_3$ 
& 58 
& 48~\textcolor{blue}{\scriptsize$\downarrow$10}   
& 94~\textcolor{blue}{\scriptsize$\uparrow$36}~\textcolor{red}{\scriptsize$\uparrow$46}
& 11 
& 7~\textcolor{blue}{\scriptsize$\downarrow$4}   
& 17~\textcolor{blue}{\scriptsize$\uparrow$6}~\textcolor{red}{\scriptsize$\uparrow$10} \\
All 
& 76 
& 60~\textcolor{blue}{\scriptsize$\downarrow$16}   
& 98~\textcolor{blue}{\scriptsize$\uparrow$22}~\textcolor{red}{\scriptsize$\uparrow$38}
& 30 
& 20~\textcolor{blue}{\scriptsize$\downarrow$10}   
& 44~\textcolor{blue}{\scriptsize$\uparrow$14}~\textcolor{red}{\scriptsize$\uparrow$24} \\
\bottomrule
\end{tabularx}
\caption{Averages of Syntax and Semantic Correctness Grouped in $C_1$, $C_2$, and $C_3$ (\%)}
\label{tab:nonterminals_results}
\end{table*}

% Syntax and semantic correctness results of LLMS for production rules grouped
\begin{table*}
\centering
\begin{tabularx}{\textwidth}{l X X X X X X}
\toprule
\textbf{Challenge Set} & \textbf{$SX_{DP}$} & \textbf{$SX_{OPF}$} & \textbf{$SX_{\textbf{\NAME}}$} & \textbf{$SE_{DP}$} & \textbf{$SE_{OPF}$} & \textbf{$SE_{\textbf{\NAME}}$} \\
% gpt-4o
\midrule
\multicolumn{7}{c}{\small\textbf{GPT-4o}} \\
\midrule
$P_1$ 
& 100 
& 100 
& 100 
& 99 
& 99
& 100~\textcolor{blue}{\scriptsize$\uparrow$1}~\textcolor{red}{\scriptsize$\uparrow$1} \\
$P_2$ 
& 100 
& 100 
& 100 
& 96 
& 95~\textcolor{blue}{\scriptsize$\downarrow$1}  
& 100~\textcolor{blue}{\scriptsize$\uparrow$4}~\textcolor{red}{\scriptsize$\uparrow$5} \\
$P_3$ 
& 81 
& 93~\textcolor{blue}{\scriptsize$\uparrow$12} 
& 89~\textcolor{blue}{\scriptsize$\uparrow$8}~\textcolor{red}{\scriptsize$\downarrow$4} 
& 62 
& 64~\textcolor{blue}{\scriptsize$\uparrow$2} 
& 82~\textcolor{blue}{\scriptsize$\uparrow$20}~\textcolor{red}{\scriptsize$\uparrow$18} \\
All 
& 93 
& 97~\textcolor{blue}{\scriptsize$\uparrow$4}   
& 96~\textcolor{blue}{\scriptsize$\uparrow$3}~\textcolor{red}{\scriptsize$\downarrow$1} 
& 84 
& 85~\textcolor{blue}{\scriptsize$\uparrow$1}   
& 93~\textcolor{blue}{\scriptsize$\uparrow$9}~\textcolor{red}{\scriptsize$\uparrow$8} \\
% gpt-3.5-turbo
\midrule
\multicolumn{7}{c}{\small\textbf{GPT-3.5-Turbo}} \\
\midrule
$P_1$ 
& 98 
& 97~\textcolor{blue}{\scriptsize$\downarrow$1}   
& 100~\textcolor{blue}{\scriptsize$\uparrow$2}~\textcolor{red}{\scriptsize$\uparrow$3}
& 69 
& 67~\textcolor{blue}{\scriptsize$\downarrow$2}   
& 93~\textcolor{blue}{\scriptsize$\uparrow$24}~\textcolor{red}{\scriptsize$\uparrow$26} \\
$P_2$ 
& 99 
& 100~\textcolor{blue}{\scriptsize$\uparrow$1}   
& 100~\textcolor{blue}{\scriptsize$\uparrow$1}
& 18 
& 22~\textcolor{blue}{\scriptsize$\uparrow$4}   
& 42~\textcolor{blue}{\scriptsize$\uparrow$24}~\textcolor{red}{\scriptsize$\uparrow$20} \\
$P_3$ 
& 86 
& 90~\textcolor{blue}{\scriptsize$\uparrow$4}   
& 96~\textcolor{blue}{\scriptsize$\uparrow$10}~\textcolor{red}{\scriptsize$\uparrow$6}
& 23 
& 24~\textcolor{blue}{\scriptsize$\uparrow$1}   
& 46~\textcolor{blue}{\scriptsize$\uparrow$23}~\textcolor{red}{\scriptsize$\uparrow$22} \\
All 
& 94 
& 95~\textcolor{blue}{\scriptsize$\uparrow$1}   
& 99~\textcolor{blue}{\scriptsize$\uparrow$5}~\textcolor{red}{\scriptsize$\uparrow$4}
& 37 
& 38~\textcolor{blue}{\scriptsize$\uparrow$1}   
& 61~\textcolor{blue}{\scriptsize$\uparrow$24}~\textcolor{red}{\scriptsize$\uparrow$23} \\
% qwen:72b-chat
\midrule
\multicolumn{7}{c}{\small\textbf{Qwen:72b-Chat}} \\
\midrule
$P_1$ 
& 72 
& 74~\textcolor{blue}{\scriptsize$\uparrow$2}   
& 96~\textcolor{blue}{\scriptsize$\uparrow$24}~\textcolor{red}{\scriptsize$\uparrow$22}
& 42 
& 43~\textcolor{blue}{\scriptsize$\uparrow$1}   
& 69~\textcolor{blue}{\scriptsize$\uparrow$27}~\textcolor{red}{\scriptsize$\uparrow$26} \\
$P_2$ 
& 51 
& 51
& 76~\textcolor{blue}{\scriptsize$\uparrow$25}~\textcolor{red}{\scriptsize$\uparrow$25}
& 12 
& 13~\textcolor{blue}{\scriptsize$\uparrow$1}   
& 24~\textcolor{blue}{\scriptsize$\uparrow$12}~\textcolor{red}{\scriptsize$\uparrow$11} \\
$P_3$ 
& 21 
& 25~\textcolor{blue}{\scriptsize$\uparrow$4}   
& 59~\textcolor{blue}{\scriptsize$\uparrow$38}~\textcolor{red}{\scriptsize$\uparrow$34}
& 6 
& 7~\textcolor{blue}{\scriptsize$\uparrow$1}   
& 21~\textcolor{blue}{\scriptsize$\uparrow$15}~\textcolor{red}{\scriptsize$\uparrow$14} \\
All 
& 47 
& 49~\textcolor{blue}{\scriptsize$\uparrow$2}   
& 76~\textcolor{blue}{\scriptsize$\uparrow$29}~\textcolor{red}{\scriptsize$\uparrow$27}
& 20 
& 21~\textcolor{blue}{\scriptsize$\uparrow$1}   
& 38~\textcolor{blue}{\scriptsize$\uparrow$18}~\textcolor{red}{\scriptsize$\uparrow$17} \\
% llama3:70b-instruct
\midrule
\multicolumn{7}{c}{\small\textbf{Llama3:70b-Instruct}} \\
\midrule
$P_1$ 
& 86 
& 90~\textcolor{blue}{\scriptsize$\uparrow$4}   
& 97~\textcolor{blue}{\scriptsize$\uparrow$11}~\textcolor{red}{\scriptsize$\uparrow$7}
& 73 
& 73
& 92~\textcolor{blue}{\scriptsize$\uparrow$19}~\textcolor{red}{\scriptsize$\uparrow$19} \\
$P_2$ 
& 63
& 69~\textcolor{blue}{\scriptsize$\uparrow$6}   
& 86~\textcolor{blue}{\scriptsize$\uparrow$23}~\textcolor{red}{\scriptsize$\uparrow$17}
& 38 
& 40~\textcolor{blue}{\scriptsize$\uparrow$2}   
& 67~\textcolor{blue}{\scriptsize$\uparrow$29}~\textcolor{red}{\scriptsize$\uparrow$27} \\
$P_3$ 
& 26 
& 30~\textcolor{blue}{\scriptsize$\uparrow$4}   
& 47~\textcolor{blue}{\scriptsize$\uparrow$21}~\textcolor{red}{\scriptsize$\uparrow$17}
& 15 
& 15
& 29~\textcolor{blue}{\scriptsize$\uparrow$14}~\textcolor{red}{\scriptsize$\uparrow$14} \\
All 
& 57 
& 61~\textcolor{blue}{\scriptsize$\uparrow$4}   
& 75~\textcolor{blue}{\scriptsize$\uparrow$18}~\textcolor{red}{\scriptsize$\uparrow$14}
& 41 
& 42~\textcolor{blue}{\scriptsize$\uparrow$1}   
& 61~\textcolor{blue}{\scriptsize$\uparrow$20}~\textcolor{red}{\scriptsize$\uparrow$19} \\
% gemma2:27b-instruct
\midrule
\multicolumn{7}{c}{\small\textbf{Gemma2:27b-Instruct}} \\
\midrule
$P_1$ 
& 99 
& 100~\textcolor{blue}{\scriptsize$\uparrow$1}   
& 100~\textcolor{blue}{\scriptsize$\uparrow$1}
& 87 
& 88~\textcolor{blue}{\scriptsize$\uparrow$1}   
& 98~\textcolor{blue}{\scriptsize$\uparrow$11}~\textcolor{red}{\scriptsize$\uparrow$10} \\
$P_2$ 
& 99 
& 99 
& 100~\textcolor{blue}{\scriptsize$\uparrow$1}~\textcolor{red}{\scriptsize$\uparrow$1}
& 48 
& 47~\textcolor{blue}{\scriptsize$\downarrow$1}
& 84~\textcolor{blue}{\scriptsize$\uparrow$36}~\textcolor{red}{\scriptsize$\uparrow$37} \\
$P_3$ 
& 77 
& 80~\textcolor{blue}{\scriptsize$\uparrow$3}   
& 94~\textcolor{blue}{\scriptsize$\uparrow$17}~\textcolor{red}{\scriptsize$\uparrow$14}
& 33 
& 37~\textcolor{blue}{\scriptsize$\uparrow$4}   
& 58~\textcolor{blue}{\scriptsize$\uparrow$25}~\textcolor{red}{\scriptsize$\uparrow$21} \\
All 
& 91 
& 92~\textcolor{blue}{\scriptsize$\uparrow$1}   
& 98~\textcolor{blue}{\scriptsize$\uparrow$7}~\textcolor{red}{\scriptsize$\uparrow$6}
& 56 
& 57~\textcolor{blue}{\scriptsize$\uparrow$1}   
& 79~\textcolor{blue}{\scriptsize$\uparrow$23}~\textcolor{red}{\scriptsize$\uparrow$22} \\
% mistral:7b-instruct
\midrule
\multicolumn{7}{c}{\small\textbf{Mistral:7b-Instruct}} \\
\midrule
$P_1$ 
& 2 
& 26~\textcolor{blue}{\scriptsize$\uparrow$24}   
& 3~\textcolor{blue}{\scriptsize$\uparrow$2}~\textcolor{red}{\scriptsize$\downarrow$23}
& 1 
& 15~\textcolor{blue}{\scriptsize$\uparrow$14}   
& 2~\textcolor{blue}{\scriptsize$\uparrow$1}~\textcolor{red}{\scriptsize$\downarrow$13} \\
$P_2$ 
& 0 
& 19~\textcolor{blue}{\scriptsize$\uparrow$19}    
& 1~\textcolor{blue}{\scriptsize$\uparrow$1}~\textcolor{red}{\scriptsize$\downarrow$18}
& 0 
& 5~\textcolor{blue}{\scriptsize$\uparrow$5}    
& 0~\textcolor{red}{\scriptsize$\downarrow$5} \\
$P_3$ 
& 0 
& 13~\textcolor{blue}{\scriptsize$\uparrow$13}   
& 0~\textcolor{red}{\scriptsize$\downarrow$13}
& 0 
& 3~\textcolor{blue}{\scriptsize$\uparrow$3}   
& 0~\textcolor{red}{\scriptsize$\downarrow$3} \\
All 
& 1 
& 19~\textcolor{blue}{\scriptsize$\uparrow$18}   
& 1~\textcolor{red}{\scriptsize$\downarrow$18}
& 0 
& 8~\textcolor{blue}{\scriptsize$\uparrow$8}   
& 1~\textcolor{blue}{\scriptsize$\uparrow$1}~\textcolor{red}{\scriptsize$\downarrow$7} \\
% codestral:22b
\midrule
\multicolumn{7}{c}{\small\textbf{Codestral:22b}} \\
\midrule
$P_1$ 
& 82 
& 97~\textcolor{blue}{\scriptsize$\uparrow$15}   
& 99~\textcolor{blue}{\scriptsize$\uparrow$17}~\textcolor{red}{\scriptsize$\uparrow$2}
& 79 
& 89~\textcolor{blue}{\scriptsize$\uparrow$10}   
& 97~\textcolor{blue}{\scriptsize$\uparrow$18}~\textcolor{red}{\scriptsize$\uparrow$8} \\
$P_2$ 
& 51 
& 73~\textcolor{blue}{\scriptsize$\uparrow$22}   
& 80~\textcolor{blue}{\scriptsize$\uparrow$29}~\textcolor{red}{\scriptsize$\uparrow$7}
& 35 
& 38~\textcolor{blue}{\scriptsize$\uparrow$3}   
& 66~\textcolor{blue}{\scriptsize$\uparrow$31}~\textcolor{red}{\scriptsize$\uparrow$28} \\
$P_3$ 
& 29 
& 47~\textcolor{blue}{\scriptsize$\uparrow$18}   
& 64~\textcolor{blue}{\scriptsize$\uparrow$35}~\textcolor{red}{\scriptsize$\uparrow$17}
& 21 
& 30~\textcolor{blue}{\scriptsize$\uparrow$9}   
& 41~\textcolor{blue}{\scriptsize$\uparrow$20}~\textcolor{red}{\scriptsize$\uparrow$11} \\
All 
& 53 
& 71~\textcolor{blue}{\scriptsize$\uparrow$18}   
& 80~\textcolor{blue}{\scriptsize$\uparrow$27}~\textcolor{red}{\scriptsize$\uparrow$9}
& 44 
& 52~\textcolor{blue}{\scriptsize$\uparrow$8}   
& 67~\textcolor{blue}{\scriptsize$\uparrow$23}~\textcolor{red}{\scriptsize$\uparrow$15} \\
% starcoder2:15b-instruct
\midrule
\multicolumn{7}{c}{\small\textbf{Starcoder2:15b-Instruct}} \\
\midrule
$P_1$ 
& 96 
& 68~\textcolor{blue}{\scriptsize$\downarrow$28}   
& 100~\textcolor{blue}{\scriptsize$\uparrow$4}~\textcolor{red}{\scriptsize$\uparrow$32}
& 54 
& 34~\textcolor{blue}{\scriptsize$\downarrow$20}   
& 74~\textcolor{blue}{\scriptsize$\uparrow$20}~\textcolor{red}{\scriptsize$\uparrow$40} \\
$P_2$ 
& 65 
& 59~\textcolor{blue}{\scriptsize$\downarrow$6}   
& 99~\textcolor{blue}{\scriptsize$\uparrow$34}~\textcolor{red}{\scriptsize$\uparrow$40}
& 16 
& 10~\textcolor{blue}{\scriptsize$\downarrow$6}   
& 26~\textcolor{blue}{\scriptsize$\uparrow$10}~\textcolor{red}{\scriptsize$\uparrow$16} \\
$P_3$ 
& 67 
& 55~\textcolor{blue}{\scriptsize$\downarrow$12}   
& 95~\textcolor{blue}{\scriptsize$\uparrow$28}~\textcolor{red}{\scriptsize$\uparrow$40}
& 20 
& 15~\textcolor{blue}{\scriptsize$\downarrow$5}   
& 31~\textcolor{blue}{\scriptsize$\uparrow$11}~\textcolor{red}{\scriptsize$\uparrow$16} \\
All 
& 76 
& 60~\textcolor{blue}{\scriptsize$\downarrow$16}   
& 98~\textcolor{blue}{\scriptsize$\uparrow$22}~\textcolor{red}{\scriptsize$\uparrow$38}
& 30 
& 20~\textcolor{blue}{\scriptsize$\downarrow$10}   
& 44~\textcolor{blue}{\scriptsize$\uparrow$14}~\textcolor{red}{\scriptsize$\uparrow$24} \\
\bottomrule
\end{tabularx}
\caption{Averages of Syntax and Semantic Correctness Grouped in $P_1$, $P_2$, and $P_3$ (\%)}
\label{tab:prs_results}
\end{table*}

% FGI Diff results of LLMS for production rules grouped
\begin{table*}
\centering
\begin{tabularx}{\textwidth}{l X X X}
\toprule
\textbf{Challenge Set} & \textbf{$\diff^\diamond_{DP}$} & \textbf{$\diff^\diamond_{OPF}$} & \textbf{$\diff^\diamond_{\textbf{\NAME}}$} \\
% gpt-4o
\midrule
\multicolumn{4}{c}{\small\textbf{GPT-4o}} \\
\midrule
$P_1$ 
& 0.22 
& 0.22 
& 0.21 \\
$P_2$ 
& 1.38 
& 1.32 
& 1.18 \\
$P_3$ 
& 4.37
& 3.95
& 3.36\\
All 
& 1.76
& 1.63 
& 1.65\\
% gpt-3.5-turbo
\midrule
\multicolumn{4}{c}{\small\textbf{GPT-3.5-Turbo}} \\
\midrule
$P_1$ 
& 0.42 
& 0.37
& 0.27\\
$P_2$ 
& 2.43 
& 2.34
& 2.21\\
$P_3$ 
& 3.38 
& 3.42
& 4.30\\
All 
& 1.40 
& 1.43
& 1.81\\
% qwen:72b-chat
\midrule
\multicolumn{4}{c}{\small\textbf{Qwen:72b-Chat}} \\
\midrule
$P_1$ 
& 0.12 
& 0.08
& -0.02\\
$P_2$ 
& 0.32 
& 0.38 
& 0.38\\
$P_3$ 
& 2.33 
& 3.00
& 3.02\\
All 
& 0.04 
& 0.50
& 0.68\\
% llama3:70b-instruct
\midrule
\multicolumn{4}{c}{\small\textbf{Llama3:70b-Instruct}} \\
\midrule
$P_1$ 
& 0.39 
& 0.38
& 0.44\\
$P_2$ 
& 1.56 
& 1.62
& 1.50\\
$P_3$ 
& 2.52 
& 2.42 
& 2.85\\
All 
& 1.00 
& 1.00 
& 1.21\\
% gemma2:27b-instruct
\midrule
\multicolumn{4}{c}{\small\textbf{Gemma2:27b-Instruct}} \\
\midrule
$P_1$ 
& 0.55 
& 0.54
& 0.59\\
$P_2$ 
& 1.77 
& 1.74 
& 1.50\\
$P_3$ 
& 4.03 
& 4.27
& 4.40\\
All 
& 1.64 
& 1.74
& 1.93\\
% mistral:7b-instruct
\midrule
\multicolumn{4}{c}{\small\textbf{Mistral:7b-Instruct}} \\
\midrule
$P_1$ 
& 1.00 
& -0.59
& 0.00\\
$P_2$ 
& N/A 
& 1.88
& N/A\\
$P_3$ 
& N/A
& 3.17 
& N/A\\
All 
& 1.00 
& 0.44
& 0.00\\
% codestral:22b
\midrule
\multicolumn{4}{c}{\small\textbf{Codestral:22b}} \\
\midrule
$P_1$ 
& 0.18 
& 0.16
& 0.10\\
$P_2$ 
& 1.26
& 1.17
& 0.75\\
$P_3$ 
& 1.86
& 2.34
& 2.96\\
All 
& 0.72 
& 0.85
& 0.94\\
% starcoder2:15b-instruct
\midrule
\multicolumn{4}{c}{\small\textbf{Starcoder2:15b-Instruct}} \\
\midrule
$P_1$ 
& 0.39
& 0.41
& 0.20\\
$P_2$ 
& 0.40 
& 0.62
& 0.71\\
$P_3$ 
& 2.90
& 3.13 
& 4.11\\
All 
& 1.02 
& 1.22
& 1.32\\
\bottomrule
\end{tabularx}
\caption{Averages of $\diff^\diamond$ Grouped in $P_1$, $P_2$, and $P_3$}
\label{tab:fgi_diff}
\end{table*}

% FGI OF results of LLMS for production rules grouped
\begin{table*}
\centering
\begin{tabularx}{\textwidth}{l X X X}
\toprule
\textbf{Challenge Set} & \textbf{$\of_{DP}$} & \textbf{$\of_{OPF}$} & \textbf{$\of_{\textbf{\NAME}}$} \\
% gpt-4o
\midrule
\multicolumn{4}{c}{\small\textbf{GPT-4o}} \\
\midrule
$P_1$ 
& 0.00 
& 0.00 
& 0.00 \\
$P_2$ 
& 4.03 
& 0.68 
& 1.28 \\
$P_3$ 
& 13.49 
& 11.54 
& 11.31\\
All 
& 5.08
& 3.50 
& 4.17\\
% gpt-3.5-turbo
\midrule
\multicolumn{4}{c}{\small\textbf{GPT-3.5-Turbo}} \\
\midrule
$P_1$ 
& 0.00 
& 0.00
& 0.00\\
$P_2$ 
& 35.71
& 34.29
& 24.24\\
$P_3$ 
& 17.02
& 16.67
& 24.47\\
All 
& 9.05 
& 9.85
& 11.89\\
% qwen:72b-chat
\midrule
\multicolumn{4}{c}{\small\textbf{Qwen:72b-Chat}} \\
\midrule
$P_1$ 
& 0.00 
& 0.00
& 0.00\\
$P_2$ 
& 0.00 
& 0.00 
& 0.00\\
$P_3$ 
& 25.00 
& 21.43
& 19.05\\
All 
& 2.8 
& 2.65
& 3.92\\
% llama3:70b-instruct
\midrule
\multicolumn{4}{c}{\small\textbf{Llama3:70b-Instruct}} \\
\midrule
$P_1$ 
& 0.00
& 0.00
& 0.00\\
$P_2$ 
& 1.67 
& 4.76 
& 0.95\\
$P_3$ 
& 6.45 
& 6.45 
& 10.17\\
All 
& 1.35 
& 2.21
& 2.12\\
% gemma2:27b-instruct
\midrule
\multicolumn{4}{c}{\small\textbf{Gemma2:27b-Instruct}} \\
\midrule
$P_1$ 
& 0.00
& 0.00
& 0.00\\
$P_2$ 
& 9.33 
& 9.59 
& 6.11\\
$P_3$ 
& 16.18
& 16.00
& 14.41\\
All 
& 6.00 
& 6.19
& 5.88\\
% mistral:7b-instruct
\midrule
\multicolumn{4}{c}{\small\textbf{Mistral:7b-Instruct}} \\
\midrule
$P_1$ 
& 0.00 
& 0.00
& 0.00\\
$P_2$ 
& N/A 
& 12.05
& N/A\\
$P_3$ 
& N/A
& 33.33
& N/A\\
All 
& 0 
& 7.32
& 0\\
% codestral:22b
\midrule
\multicolumn{4}{c}{\small\textbf{Codestral:22b}} \\
\midrule
$P_1$ 
& 0.00
& 0.00
& 0.00\\
$P_2$ 
& 0.00
& 0.00
& 0.00\\
$P_3$ 
& 4.76
& 6.56
& 10.84\\
All 
& 0.84 
& 1.42
& 2.49\\
% starcoder2:15b-instruct
\midrule
\multicolumn{4}{c}{\small\textbf{Starcoder2:15b-Instruct}} \\
\midrule
$P_1$ 
& 0.00
& 0.00
& 0.00\\
$P_2$ 
& 16.00 
& 6.25
& 4.88\\
$P_3$ 
& 12.20 
& 12.90 
& 15.87\\
All 
& 5.49 
& 4.63
& 5.04\\
\bottomrule
\end{tabularx}
\caption{Averages of $\of$ Grouped in $P_1$, $P_2$, and $P_3$ (\%)}
\label{tab:fgi_of}
\end{table*}

% FGI OG results of LLMS for production rules grouped
\begin{table*}
\centering
\begin{tabularx}{\textwidth}{l X X X}
\toprule
\textbf{Challenge Set} & \textbf{$\og_{DP}$} & \textbf{$\og_{OPF}$} & \textbf{$\og_{\textbf{\NAME}}$} \\
% gpt-4o
\midrule
\multicolumn{4}{c}{\small\textbf{GPT-4o}} \\
\midrule
$P_1$ 
& 0.00 
& 0.00 
& 1.11 \\
$P_2$ 
& 0.00 
& 0.00 
& 0.00 \\
$P_3$ 
& 0.00 
& 0.00
& 0.60\\
All 
& 0.00
& 0.00 
& 0.60\\
% gpt-3.5-turbo
\midrule
\multicolumn{4}{c}{\small\textbf{GPT-3.5-Turbo}} \\
\midrule
$P_1$ 
& 0.00
& 0.00
& 1.19\\
$P_2$ 
& 0.00 
& 0.00 
& 0.00\\
$P_3$ 
& 2.13 
& 0.00 
& 1.06\\
All 
& 0.50 
& 0.00 
& 0.91\\
% qwen:72b-chat
\midrule
\multicolumn{4}{c}{\small\textbf{Qwen:72b-Chat}} \\
\midrule
$P_1$ 
& 1.32 
& 2.56
& 3.20\\
$P_2$ 
& 0.00 
& 0.00 
& 2.70 \\
$P_3$ 
& 8.33
& 0.00
& 0.00\\
All 
& 1.87 
& 1.77
& 2.45\\
% llama3:70b-instruct
\midrule
\multicolumn{4}{c}{\small\textbf{Llama3:70b-Instruct}} \\
\midrule
$P_1$ 
& 0.00 
& 0.00
& 0.00\\
$P_2$ 
& 0.00 
& 0.00
& 0.00\\
$P_3$ 
& 0.00 
& 0.00
& 0.00\\
All 
& 0.00 
& 0.00
& 0.00\\
% gemma2:27b-instruct
\midrule
\multicolumn{4}{c}{\small\textbf{Gemma2:27b-Instruct}} \\
\midrule
$P_1$ 
& 0.00 
& 0.00
& 0.00\\
$P_2$ 
& 0.00 
& 1.37 
& 1.53\\
$P_3$ 
& 0.00 
& 0.00 
& 0.00\\
All 
& 0.00
& 0.33
& 0.47\\
% mistral:7b-instruct
\midrule
\multicolumn{4}{c}{\small\textbf{Mistral:7b-Instruct}} \\
\midrule
$P_1$ 
& 0.00 
& 11.11
& 0.00\\
$P_2$ 
& N/A 
& 0.00
& N/A\\
$P_3$ 
& N/A
& 0.00
& N/A\\
All 
& 0.00 
& 7.32
& 0.00\\
% codestral:22b
\midrule
\multicolumn{4}{c}{\small\textbf{Codestral:22b}} \\
\midrule
$P_1$ 
& 0.70
& 1.25
& 1.14\\
$P_2$ 
& 1.85
& 1.67
& 0.97\\
$P_3$ 
& 0.00 
& 0.00
& 0.00\\
All 
& 0.84 
& 1.07
& 0.83\\
% starcoder2:15b-instruct
\midrule
\multicolumn{4}{c}{\small\textbf{Starcoder2:15b-Instruct}} \\
\midrule
$P_1$ 
& 0.00 
& 0.00 
& 2.24\\
$P_2$ 
& 12.00 
& 0.00 
& 4.88\\
$P_3$ 
& 0.00 
& 0.00 
& 0.00\\
All 
& 1.83
& 0.00 
& 2.10\\
\bottomrule
\end{tabularx}
\caption{Averages of $\og$ Grouped in $P_1$, $P_2$, and $P_3$ (\%)}
\label{tab:fgi_og}
\end{table*}

% TU results of LLMS for production rules grouped
\begin{table*}
\centering
\begin{tabularx}{\textwidth}{l X X X}
\toprule
\textbf{Challenge Set} & \textbf{$\tu^\diamond_{DP}$} & \textbf{$\tu^\diamond_{OPF}$} & \textbf{$\og^\diamond_{\textbf{\NAME}}$} \\
% gpt-4o
\midrule
\multicolumn{4}{c}{\small\textbf{GPT-4o}} \\
\midrule
$P_1$ 
& 100 
& 100 
& 99.81 \\
$P_2$ 
& 99.60 
& 99.55 
& 99.67 \\
$P_3$ 
& 93.04 
& 92.80
& 91.39\\
All 
& 97.93 
& 97.81 
& 96.96 \\
% gpt-3.5-turbo
\midrule
\multicolumn{4}{c}{\small\textbf{GPT-3.5-Turbo}} \\
\midrule
$P_1$ 
& 99.64
& 99.33
& 97.93\\
$P_2$ 
& 90.91 
& 89.71 
& 89.11\\
$P_3$ 
& 78.56 
& 79.74 
& 77.57\\
All 
& 93.43 
& 93.04 
& 90.32\\
% qwen:72b-chat
\midrule
\multicolumn{4}{c}{\small\textbf{Qwen:72b-Chat}} \\
\midrule
$P_1$ 
& 92.95 
& 94.66
& 90.47\\
$P_2$ 
& 76.56
& 74.29 
& 81.31 \\
$P_3$ 
& 79.79
& 77.88
& 77.53\\
All 
& 88.56
& 88.79
& 86.14\\
% llama3:70b-instruct
\midrule
\multicolumn{4}{c}{\small\textbf{Llama3:70b-Instruct}} \\
\midrule
$P_1$ 
& 100 
& 100
& 99.80\\
$P_2$ 
& 88.48 
& 89.06
& 93.03\\
$P_3$ 
& 97.73
& 98.24
& 87.33\\
All 
& 96.58
& 96.71
& 95.41\\
% gemma2:27b-instruct
\midrule
\multicolumn{4}{c}{\small\textbf{Gemma2:27b-Instruct}} \\
\midrule
$P_1$ 
& 99.82 
& 99.82 
& 99.25\\
$P_2$ 
& 92.67 
& 94.21 
& 95.80\\
$P_3$ 
& 93.06
& 94.02
& 90.61\\
All 
& 96.50
& 97.07
& 95.78\\
% mistral:7b-instruct
\midrule
\multicolumn{4}{c}{\small\textbf{Mistral:7b-Instruct}} \\
\midrule
$P_1$ 
& 50.00 
& 68.34
& 83.33\\
$P_2$ 
& N/A 
& 64.67
& N/A\\
$P_3$ 
& N/A
& 64.17
& N/A\\
All 
& 50.00 
& 67.01
& 83.33\\
% codestral:22b
\midrule
\multicolumn{4}{c}{\small\textbf{Codestral:22b}} \\
\midrule
$P_1$ 
& 98.83
& 98.11
& 98.36\\
$P_2$ 
& 88.76
& 90.31
& 89.58\\
$P_3$ 
& 87.33
& 85.27
& 83.37\\
All 
& 94.54
& 93.66
& 92.41\\
% starcoder2:15b-instruct
\midrule
\multicolumn{4}{c}{\small\textbf{Starcoder2:15b-Instruct}} \\
\midrule
$P_1$ 
& 97.86 
& 96.44 
& 96.64\\
$P_2$ 
& 83.29 
& 87.57 
& 81.92\\
$P_3$ 
& 84.91 
& 85.39
& 80.48\\
All 
& 92.40
& 91.96 
& 89.83\\
\bottomrule
\end{tabularx}
\caption{Averages of $TU^\diamond$ Grouped in $P_1$, $P_2$, and $P_3$ (\%)}
\label{tab:fgi_tu}
\end{table*}

\end{document}